\documentclass[a4paper,fleqn]{cas-sc}

\usepackage[authoryear,longnamesfirst]{natbib}

\usepackage{amsmath,amssymb,bm}

\usepackage{graphicx}
\usepackage{booktabs,tabularx,multirow}
\usepackage{subcaption}

\usepackage{tikz}
\usetikzlibrary{arrows.meta,positioning}

\usepackage{placeins}

\usepackage[english]{babel}
\usepackage{float}

\begin{document}

% Running title and author names
\shorttitle{Selective Transfer Learning for Groundwater Forecasting}
\shortauthors{Wu et al.}

% Article title
\title[mode=title]{Selective Transfer Learning for Groundwater
Level Anomaly Forecasting: Accuracy--Risk Trade-offs}

% Authors
\author[1]{Yaxuan Wu}
\ead{stephanie.wu2@outlook.com}

\author[1]{Ali Mohammed Mansoor Alsahag}
\cormark[1]
\ead{a.m.m.alsahag@uva.nl}

\author[1]{Seyed Sahand Mohammadi Ziabari}
\ead{s.s.mohammadiziabari@uva.nl}

% Affiliation
\affiliation[1]{
    organization={Informatics Institute, University of Amsterdam},
    city={Amsterdam},
    country={The Netherlands}
}

% Corresponding author
\cortext[1]{Corresponding author}

\begin{abstract}
This work studies context-aware selective transfer learning for
multi-well groundwater anomaly forecasting in the Netherlands.
Source-domain pretraining can raise average accuracy but is not
uniformly reliable: it degrades performance on a substantial fraction
of heterogeneous wells. Context-Aware Selective Learning (CASL) uses
calibration-trained, constrained contextual blending of frozen
no-transfer and fixed-transfer parents. Global Convex Blend uses the
same parents with one calibration-fitted weight per seed. The
corrected task requires nine consecutive observed monthly inputs and
an exact three-calendar-month target, without interpolation.
Evaluation covers 457 wells and 13{,}718 test windows under five
predefined seeds, with the four-cell \texttt{maxsim0.60} source pool
frozen using Always-TL validation RMSE alone. Fixed transfer harms
$18.38$--$51.86\%$ of wells. CASL has $0.424$--$1.084\%$ lower RMSE
than No TL, partially rescuing $97.53$--$99.50\%$ and fully rescuing
$23.46$--$43.78\%$ of fixed-harmed wells. Partial rescue can arise
structurally from convex blending and does not establish contextual
superiority. Global Convex Blend protects more wells in four seeds,
while CASL has slightly lower pooled RMSE in four. Neither dominates:
the results support an accuracy--risk trade-off on the fixed split,
not rolling deployment validity.
\end{abstract}


\maketitle

% Main manuscript sections
\section{Introduction}
\label{sec:introduction}

Groundwater is a critical resource for water security and climate adaptation. In low-lying delta regions such as the Netherlands, groundwater fluctuations affect freshwater availability, subsurface infrastructure, ecosystem functioning, and drought preparedness, so reliable short-range forecasts matter for operational planning and early warning under growing climate variability and human pressure \cite{Deltaprogramma2024Protecting2024,greenSurfaceGlobalChange2011,vanloonDroughtAnthropocene2016}. Across the Netherlands, monitoring conditions vary widely --- from sandy infiltration areas in the east, to peat and clay polders in the west, to densely managed urban environments --- so a national forecasting system must cope with both broad hydroclimatic forcing and well-specific local context.

This study addresses the following problem: given monthly hydro-meteorological anomalies and site context for all eligible Dutch BRO wells under the 1996--2016 modelling constraints --- a corrected Netherlands target dataset of 457 wells and 13{,}718 site-level test windows --- forecast standardised groundwater level anomalies relative to NAP ($wtda$) at exactly $H = 3$ calendar months, using nine consecutive observed monthly inputs without interpolation, and determine \emph{when} transfer learning improves forecasting performance rather than harms it. Monthly resolution suits groundwater, which responds more slowly than surface water, and matches anomaly-based groundwater drought monitoring in the hydrological literature \cite{bloomfieldAnalysisGroundwaterDrought2013a,kumarMultiscaleEvaluationStandardized2016}; the 3-month lead is operationally relevant for drought preparedness, seasonal freshwater planning, and short-term management \cite{Deltaprogramma2024Protecting2024,profetaPredictingGroundwaterLevel2025a}. The target is $wtda$ rather than raw groundwater levels because anomaly-based targets improve comparability across wells with different baselines and better reflect drought-relevant deviations than site-specific level offsets \cite{maUsingLongShortTerm2021,maAdvancingAIbasedPanEuropean2022,bloomfieldAnalysisGroundwaterDrought2013a}.

Machine learning has substantially improved groundwater forecasting compared with purely classical statistical and physically based approaches, especially when complex nonlinear dynamics and multiple forcing variables are involved \cite{rajaeeReviewArtificialIntelligence2019,booGroundwaterLevelForecasting2024,wunschGroundwaterLevelForecasting2021}. Among deep-learning methods, LSTM is a widely used baseline for modelling temporal dependencies in hydrological time series \cite{hochreiterLongShortTermMemory1997}. At continental scale, Ma et al.\ showed that transfer learning (TL) can transfer hydrometeorological knowledge from simulated European-scale data to observational groundwater anomaly prediction, and that adding soil-moisture anomalies to precipitation improves $wtda$ estimation \cite{maAdvancingAIbasedPanEuropean2022,maIndirectApproachBased2021}. At Dutch regional scale, Profeta et al.\ adapted this LSTM--TL framework to Zuid-Holland and reported improved $wtda$ forecasting relative to a non-transfer LSTM baseline \cite{profetaPredictingGroundwaterLevel2025a}. However, prior groundwater TL studies mainly evaluate whether transfer improves performance \emph{on average} and leave a more practically important question insufficiently tested: \emph{is transfer learning beneficial for every target well?} In heterogeneous, human-influenced groundwater systems, wells differ substantially in hydrogeology, forcing-response relationships, management effects, and data availability, so transferred representations may help some wells but degrade others, leading to negative transfer \cite{zhangSurveyNegativeTransfer2023}. This is especially relevant in national-scale multi-well forecasting, where a single modelling strategy is often assumed to generalise across wells despite substantial local variation.

Accordingly, this study reframes the problem from ``does TL help?'' to ``\emph{when} does TL help, and can it be applied more selectively?'' The LSTM--TL pipeline of Profeta et al.\ informs the modelling design. Four main policies are evaluated under identical preprocessing, chronological splits and input windows: \emph{No TL}, fixed transfer through the Deep Context LSTM (Always TL) parent, \emph{Global Convex Blend}, and \emph{Context-Aware Selective Learning (CASL)}. The parent models use matched target-training settings. Global Convex Blend uses the same frozen parents with one calibration-fitted weight per seed as a non-contextual comparison, while CASL learns a bounded per-observation contextual blend. A Site-Hard Expert Selector makes a discrete site-level choice as a secondary comparison. The source-filtering rule, \texttt{maxsim0.60}, was selected once using corrected Always-TL validation RMSE alone and retains four of 512 source cells. The filter is frozen before gate fitting and across seeds; CASL scores, calibration labels and test labels do not select it. This comparison distinguishes pooled predictive accuracy from well-level transfer harm without assuming that contextual adaptation uniquely provides protection.

The project began as an Amsterdam case study that motivated the selective-transfer design, but the final reported evidence uses the source-compatible Netherlands target panel, so Amsterdam is only the methodological starting point rather than the empirical scope.

Source--target similarity motivates source filtering and serves as an exploratory descriptor of residual CASL harm. The per-well fixed-transfer effect is defined as $\Delta_i^{\text{TL}} = \text{RMSE}_i(\text{Always TL}) - \text{RMSE}_i(\text{No TL})$. The main formal hypothesis is \textbf{Main hypothesis (selective transfer hypothesis):} a context-aware selective TL strategy should reduce the negative-transfer rate (the proportion of wells with $\Delta_i^{\text{Selective}} > 0$, where $\Delta_i^{\text{Selective}} = \text{RMSE}_i(\text{CASL}) - \text{RMSE}_i(\text{No TL})$) relative to the fixed-transfer Deep Context LSTM (Always TL) parent, while matching or improving pooled RMSE relative to No TL and Always TL under the same evaluation protocol.

This study examines the extent to which transfer learning is
conditionally beneficial for multi-well groundwater anomaly
forecasting in the Netherlands, and whether a context-aware
selective transfer strategy can reduce negative transfer while
matching or improving no-transfer and fixed-transfer baselines
under a fair evaluation protocol.

The main contribution of this study is a decision framework for
transfer learning in multi-well groundwater forecasting,
evaluated at both the pooled and the per-well level. The framework
uses separate shared-hydroclimatic and local-context encoders
and trains a calibration-stage selector on top of frozen
no-transfer and fixed-transfer parents. A matched empirical
comparison of no transfer, fixed transfer, global convex blending
and selective transfer is conducted on the corrected Netherlands
target dataset, using a frozen validation-selected source filter
(\texttt{maxsim0.60}) and five predefined seeds.

CASL remains competitive in aggregate accuracy, with
$0.424$--$1.084\%$ lower test RMSE than No TL across the five
seeds, while fixed transfer harms $18.38$--$51.86\%$ of wells.
CASL partially rescues $97.53$--$99.50\%$ and fully rescues
$23.46$--$43.78\%$ of fixed-harmed wells, but partial rescue can
also arise structurally from convex blending toward No TL.
Global Convex Blend generally protects more wells and CASL
more often gives slightly lower pooled RMSE, so neither
dominates.

The study also examines heterogeneous outcomes across seeds:
recent-state and retrospective-shift descriptors show repeated
associations, while source--target similarity associations with
residual CASL harm are weak and sign-variable. These diagnostics
do not establish causal mechanisms or prospective failure
prediction. Detailed robustness checks and descriptive
diagnostics are reported in the main text where they affect
the central conclusion, and in
Appendices~\ref{sec:apx:robustness}--\ref{sec:apx:error_analysis}.

\section{Related Work}
\label{sec:related_work}

This study sits at the intersection of groundwater forecasting, transfer learning, and decision-oriented model evaluation. Rather than treating transfer learning (TL) as universally beneficial, it focuses on a more specific question that remains underexplored in groundwater forecasting: \emph{under what conditions does transfer help, and when can it hurt?} The relevant literature is therefore organised into four connected areas: groundwater forecasting and TL baselines, conditional transfer and negative-transfer evidence, multi-site heterogeneity and architectural choices, and drought-relevant evaluation.

\textbf{Groundwater forecasting and transfer learning.} Machine learning, in particular LSTM models, has become a standard approach for groundwater level and anomaly forecasting and frequently outperforms classical statistical baselines when appropriate context variables and validation designs are used \cite{rajaeeReviewArtificialIntelligence2019,booGroundwaterLevelForecasting2024,wunschGroundwaterLevelForecasting2021,hochreiterLongShortTermMemory1997}. A central challenge in groundwater is that monitoring records are sparse, uneven across wells, and heterogeneous in both data quality and hydrogeological behaviour. This motivates methods that can transfer information from data-rich settings to data-poor sites.

At pan-European scale, Ma et al.\ map TSMP-derived hydrometeorological anomalies to groundwater table depth anomalies and use transfer learning from physically consistent simulation products to observational targets \cite{maUsingLongShortTerm2021,maIndirectApproachBased2021,maAdvancingAIbasedPanEuropean2022}. Their results show that combining precipitation anomalies with soil-moisture anomalies improves $wtda$ estimation compared with precipitation-only inputs, and that pretrained representations can transfer from simulation to observation. At Dutch regional scale, Profeta et al.\ adapted this pipeline to Zuid-Holland and reported improved $wtda$ forecasting under partial freezing and fine-tuning of the LSTM encoder \cite{profetaPredictingGroundwaterLevel2025a}. This earlier Dutch regional LSTM--TL pipeline is the closest comparison protocol for the present study; its baseline variants belong to the earlier uncorrected benchmark and were not rerun for the corrected five-seed comparison. Both lines of work evaluate transfer mainly through average gains over a non-transfer baseline; neither tests whether transfer should be applied conditionally at the level of individual wells, nor proposes a practical decision rule for reducing negative transfer in heterogeneous groundwater settings.

\textbf{Conditional transfer and negative transfer.} Transfer learning broadly relaxes the assumption that the source and target data come from the same distribution and feature space, with domain adaptation as a closely related setting \cite{panSurveyTransferLearning2010}. The broader transfer-learning literature makes clear that transfer is not guaranteed to be beneficial. When the source and target domains are insufficiently aligned, transferred representations may degrade rather than improve performance, a phenomenon commonly referred to as negative transfer \cite{zhangSurveyNegativeTransfer2023}. Two responses have emerged outside groundwater. Transferability estimation tries to assess whether pretrained representations are likely to help a downstream task before full deployment \cite{huangFrustratinglyEasyTransferability2021}, and time-series domain adaptation tries to learn what is shared across domains while retaining target-specific structure \cite{jinDomainAdaptationTime2021}. Both lines suggest that the usefulness of transfer is conditional rather than universal, and they motivate the choice in this study to design a transparent context-aware selector over a similarity-filtered source pool rather than to apply transfer by default. The selector proposed here is a simpler calibration-stage blending rule, not an equivalent of these modern transferability-estimation methods.

Inside groundwater, the same picture is emerging less explicitly. Patra et al.\ compare global and local LSTM models for regional groundwater forecasting and show that transfer and cross-site learning can be beneficial overall but vary across locations, with performance deteriorating in hydrogeologically distinct areas such as coastal wells \cite{patraRegionalGroundwaterSequential2023}. The existing selective-transfer designs in the broader literature are typically representation-based or methodologically complex, and they are designed for generic machine-learning benchmarks rather than operational groundwater forecasting. In contrast, the present study uses a deliberately simple and transparent selector: a similarity-filtered source pool together with a context-aware calibration-trained gate that learns a bounded per-observation blend between a no-transfer and a fixed-transfer parent, plus a conservative site-level expert choice as a lower-complexity comparison. CASL is constrained post-hoc blending over frozen parents, related to stacking \cite{wolpertStackedGeneralization1992} and gating/mixture-of-experts methods \cite{jacobsAdaptiveMixturesLocal1991}, rather than a new general mixture-of-experts architecture or an optimal gate. Global Convex Blend provides a non-contextual comparison, so reduced harm is not attributed to contextual adaptation alone. Multi-site forecasting therefore cannot assume that a single transfer rule will be uniformly appropriate, which directly motivates a per-well evaluation rather than only an aggregate one.

\textbf{Multi-site heterogeneity and architectural choices.} Many groundwater forecasting studies use local models trained independently for each well. Local models can perform well when long and clean records are available, but they do not exploit shared regional structure and can be brittle for short or irregular time series \cite{clarkForecastingMultipleGroundwater2022,patraRegionalGroundwaterSequential2023}. Global and multi-site models pool information across wells and can leverage common temporal patterns; recent national-scale and global studies demonstrate that this can achieve useful seasonal forecasting skill while also revealing substantial variation in performance across wells and hydrogeological settings \cite{kunzGlobalSpatialMachine2025}. Spatial structure can also be modelled explicitly through graph neural networks and spatiotemporal GCN hybrids \cite{taccariSpatialtemporalGraphNeural2024,wangGroundwaterLevelSpatiotemporal2024}, although these introduce stronger assumptions about graph construction and require neighbouring observations at prediction time. Spatial information is incorporated conservatively as static spatial context (normalised coordinates and a pumping proxy), so that the comparison between transfer policies is not confounded by spatiotemporal coupling.

Groundwater drought is increasingly understood as a coupled human--water process \cite{vanloonDroughtAnthropocene2016}; hybrid groundwater models that combine physical and data-driven components often include pumping and management variables, with anthropogenic covariates improving realism and predictive skill even when only approximate measurements are available \cite{atteaIntegrationMODFLOWDeep2025}. Regarding backbones, LSTM remains a strong baseline in hydrologic time-series modelling \cite{hochreiterLongShortTermMemory1997,wunschGroundwaterLevelForecasting2021}, and recent hydrologic benchmarking shows that it stays strong in simpler tasks while attention-based architectures become more competitive in more complex scenarios \cite{liuRNNsTransformersBenchmarking2025}; groundwater-specific Transformer applications are also emerging \cite{nieTimeSeriesWorth2022,zhouGroundwaterLevelEstimation2025}. The present study does not benchmark these stronger backbones; its question is whether a transfer-decision mechanism adds value on a controlled, moderate-capacity LSTM family, leaving the stronger-backbone test as future work.

\textbf{Drought-relevant evaluation.} Standardized groundwater anomaly indices, in particular the Standardized Groundwater level Index (SGI), support drought characterisation across heterogeneous sites \cite{bloomfieldAnalysisGroundwaterDrought2013a,bloomfieldChangesGroundwaterDrought2019}. Multi-scale standardized indices further show that groundwater drought characteristics shift under anthropogenic warming and that precipitation-only proxies such as SPI are often insufficient to capture groundwater drought reliably across locations and time scales \cite{kumarMultiscaleEvaluationStandardized2016,mckeeRELATIONSHIPDROUGHTFREQUENCY1993}. Anomaly-based targets like $wtda$ are therefore both common in the literature and operationally meaningful for drought-relevant monitoring, which is why $wtda$ is used rather than raw groundwater levels.

\textbf{Research gap.} The missing evidence in current groundwater transfer-learning work is a practical, well-level decision rule that reduces negative transfer under heterogeneous groundwater conditions inside a fair experimental protocol. Existing studies show that transfer learning can improve groundwater anomaly forecasting on average \cite{maAdvancingAIbasedPanEuropean2022,profetaPredictingGroundwaterLevel2025a}, while the broader transfer-learning literature and emerging groundwater evidence suggest that transfer can also fail when source and target are insufficiently aligned \cite{zhangSurveyNegativeTransfer2023,huangFrustratinglyEasyTransferability2021,patraRegionalGroundwaterSequential2023}. This study addresses the gap with a matched comparison of no transfer, fixed transfer, global convex blending and a similarity-filtered, context-aware selective transfer strategy on the Netherlands target dataset --- all eligible Dutch BRO wells under the 1996--2016 modelling constraints --- audited at the well level rather than only on pooled metrics, with pooled forecasting accuracy and well-level harm evaluated as distinct objectives.

\section{Methodology}
\label{sec:methodology}

\subsection{Research Design}

This study uses an empirical forecasting design to test whether a context-aware selective transfer strategy can match or improve pooled accuracy while reducing well-level negative-transfer risk relative to fixed transfer. The main comparison comprises No TL, Always TL, Global Convex Blend and CASL on the Netherlands target dataset under matched preprocessing, chronological splits, input windows and parent-training conditions. Site-Hard supplies a secondary routing comparison. Reusing the same frozen parent predictions distinguishes the transfer-policy comparison from differences in data construction or parent architecture. The design extends groundwater forecasting work by treating transfer as a well-level decision problem rather than a fixed modelling rule.

\subsection{Context-Aware Selective Learning Framework}

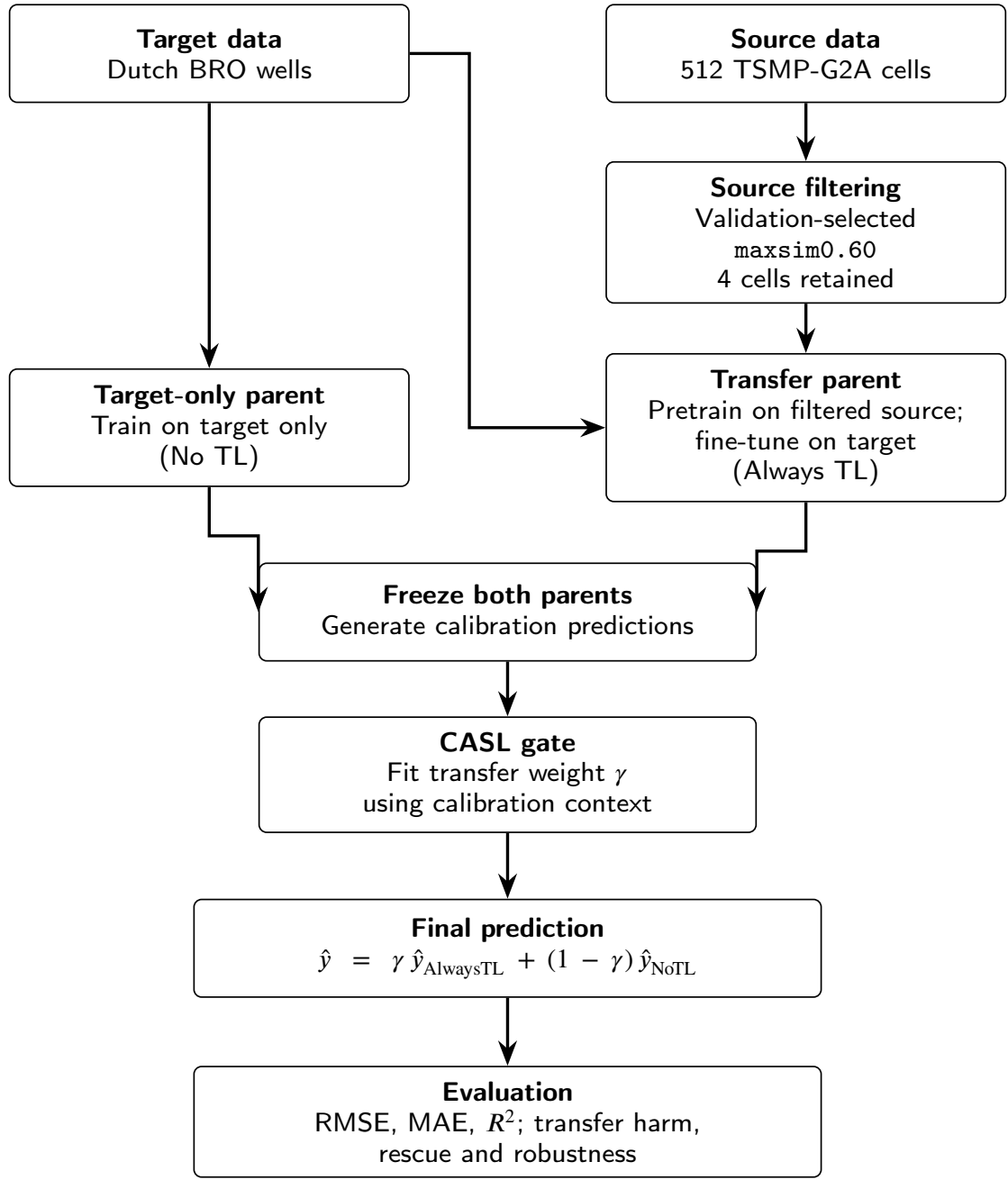
\begin{figure}[!htbp]
\centering
\resizebox{0.9\linewidth}{!}{%
\begin{tikzpicture}[
  font=\scriptsize,
  >=Stealth,
  box/.style={
    draw=black,
    fill=white,
    rectangle,
    rounded corners=2pt,
    align=center,
    text width=34mm,
    minimum height=9mm,
    inner sep=4pt
  },
  arrow/.style={->, thick},
  node distance=8mm and 12mm
]

% Top row: input data
\node[box] (target) at (0,0)
  {\textbf{Target data}\\Dutch BRO wells};

\node[box] (source) at (5.5,0)
  {\textbf{Source data}\\512 TSMP-G2A cells};

% Source filtering
\node[box] (filter) at (5.5,-1.65)
  {\textbf{Source filtering}\\
   Validation-selected \texttt{maxsim0.60}\\
   4 cells retained};

% Two parent models
\node[box] (notl) at (0,-3.45)
  {\textbf{Target-only parent}\\
   Train on target only\\
   (No TL)};

\node[box] (tl) at (5.5,-3.45)
  {\textbf{Transfer parent}\\
   Pretrain on filtered source;\\
   fine-tune on target\\
   (Always TL)};

% Calibration predictions
\node[box, text width=43mm] (freeze) at (2.75,-5.15)
  {\textbf{Freeze both parents}\\
   Generate calibration predictions};

% CASL gate
\node[box, text width=43mm] (gate) at (2.75,-6.65)
  {\textbf{CASL gate}\\
   Fit transfer weight $\gamma$\\
   using calibration context};

% Final prediction
\node[box, text width=55mm] (final) at (2.75,-8.25)
  {\textbf{Final prediction}\\
   $\hat{y} =
   \gamma\,\hat{y}_{\mathrm{AlwaysTL}} +
   (1-\gamma)\,\hat{y}_{\mathrm{NoTL}}$};
% Evaluation
\node[box, text width=55mm] (eval) at (2.75,-9.85)
  {\textbf{Evaluation}\\
   RMSE, MAE, $R^2$; transfer harm,\\
   rescue and robustness};

% Arrows: data and source filtering
\draw[arrow] (source) -- (filter);
\draw[arrow] (filter.south) -- (tl.north);

% Target data to both parent models
\draw[arrow] (target.south) -- (notl.north);
\draw[arrow] (target.east) -- ++(0.55,0)
  |- (tl.west);

% Parent predictions converge
\draw[arrow] (notl.south) -- ++(0,-0.45)
  -| (freeze.west);

\draw[arrow] (tl.south) -- ++(0,-0.45)
  -| (freeze.east);

% Main workflow
\draw[arrow] (freeze) -- (gate);
\draw[arrow] (gate) -- (final);
\draw[arrow] (final) -- (eval);

\end{tikzpicture}
}
\caption{Pipeline schematic for Context-Aware Selective Learning (CASL).}
\label{fig:casl_pipeline}
\end{figure}

The proposed framework, \textit{Context-Aware Selective Learning}, is designed for a setting in which a broader source domain is available but transfer is not expected to help all target wells equally. It combines transfer learning with an explicit selector that decides how much transfer should be used for each target case. The framework is implemented in two stages. In the first stage, two deep parent experts are trained: a target-only parent that defines a no-transfer baseline, and a transfer-learning parent whose transferable encoder is pretrained on the source domain and fine-tuned on the Netherlands target dataset. In the second stage, a selector is trained on top of the fixed parents to produce a per-observation or per-site transfer decision. This separation lets the transfer decision be learned after the parent models are fixed rather than blending model fitting and model selection into one opaque process. Figure~\ref{fig:casl_pipeline} summarises the workflow, from source filtering and parent-model training to calibration-stage blending and evaluation.

\subsubsection{Architecture and Feature Split}

A central design choice is to divide the feature space into a \emph{transferable} branch and a \emph{local context} branch. The transferable branch contains variables that exist in both the source and target domains: \texttt{tsmp\_wtda}, \texttt{pr\_a}, \texttt{sm\_a}, one-to-three-calendar-month lags for each variable, and sine/cosine month encoding. These features define the part of the model that can be pretrained on the source domain. The local context branch contains site-specific variables: the local groundwater anomaly \texttt{wtda}, the pumping proxy \texttt{pumping\_m3\_month\_log1p}, and the normalised coordinates \texttt{lon\_norm} and \texttt{lat\_norm}. Including these features prevents a purely transferable model from ignoring site-specific local context, while still allowing the shared branch to benefit from source-domain pretraining. The nine-step sequence requires consecutive observed months $t-8,\ldots,t$; the explicit climate lags extend predictor provenance to $t-11$.

The main context-aware parent uses two separate LSTM sequence encoders: a two-layer shared encoder with hidden size 64 and a one-layer local-context encoder with hidden size 28. Only the shared encoder is used in source pretraining. Their representations are concatenated with a learned 12-dimensional site embedding and the latest local groundwater state before a prediction head with 105 inputs, 64 ReLU units, dropout and a scalar output produces the final forecast. This design allows shared and local information to enter through separate branches. The two parent predictions construct calibration blend targets; predictions and their disagreement are not selector input features.

\subsubsection{Transfer Learning and Source Filtering}

The \textit{No TL} parent is the target-trained deep context LSTM without source encoder pretraining. The \textit{Always TL} parent is the same architecture whose transferable encoder is first pretrained on the source domain and then fine-tuned on the Netherlands target dataset. Because both parents share the same overall architecture, their comparison focuses on the effect of transfer rather than on a change in model family.

Before pretraining, source--target similarity diagnostics are constructed using corrected eligible target training-window origins only. For each Netherlands target site and each source cell, overlapping monthly trajectories are aligned and cosine similarity is computed on the joint signal of \texttt{wtda}/\texttt{tsmp\_wtda}, precipitation anomaly, and soil-moisture anomaly. A source cell is retained when its maximum similarity across target wells meets the candidate threshold. The four candidates are \texttt{all}, \texttt{maxsim0.40}, \texttt{maxsim0.50} and \texttt{maxsim0.60}. At seed 42, corrected Always-TL validation RMSE alone selects \texttt{maxsim0.60}, retaining four of 512 source cells; an exact tie would retain the first candidate in the listed order. Its gap to \texttt{maxsim0.50} is only $0.001150$ RMSE, so the leading candidates are numerically near-equivalent rather than evidence of an optimal threshold. The pool is frozen before calibration gate fitting and across seeds. CASL metrics, calibration labels and test labels do not select the source filter. The validation scores and retained source-cell counts for all
four candidates are reported in
Table~\ref{tab:apx_source_filter}
(Appendix~\ref{sec:apx:methodology_details}).

\subsubsection{Selector Design}

The selector supplies the calibration-stage transfer policy. It is trained on the calibration split after the parents are fixed, so the final transfer decision is not tuned on the test period and does not use test labels. Its 47 inputs are recent target-state information (the latest \texttt{wtda} and shared inputs, including calendar lags), seasonal context (sine/cosine month), site-history summaries (training-period \texttt{wtda} mean and standard deviation, pumping variability), and source--target similarity summaries (e.g.\ best-match similarity, top-$k$ mean similarity). Test-truth-derived shift descriptors are retrospective diagnostics and are not selector inputs. Recent-state and retrospective-shift descriptors later show repeated descriptive associations, while similarity associations are weak and sign-variable; this does not establish relative predictive importance or causal mechanisms. The selector learns a transfer weight for each observation, and its intended purpose is negative-transfer mitigation rather than chasing a marginal pooled-accuracy gain.

The main soft selector, the Context-Aware Selective Learning (CASL) gate, is a calibration-trained regressor whose output is clipped to a bounded blending weight $\gamma_{i,t} \in [0,1]$ between the two frozen parent predictions. Both predictions refer to the same response exactly three calendar months ahead. Its forecast is
\[
\hat{y}_{i,t}^{\text{CASL}} =
\gamma_{i,t}\hat{y}_{i,t}^{\text{AlwaysTL}} +
(1-\gamma_{i,t})\hat{y}_{i,t}^{\text{NoTL}}.
\]
The gate fits calibration-only clipped blend targets by weighted regression,
\[
L_{\text{gate}} = \frac{1}{N}\sum_{i,t}(d_{i,t}^2+0.05)(\gamma_{i,t}-\gamma_{i,t}^{*})^2 ,
\]
where $d_{i,t}$ is the Always-TL prediction minus the No-TL prediction and $\gamma_{i,t}^{*}$ is the clipped ratio of observed target minus No-TL prediction to $d_{i,t}^{\mathrm{safe}}$. The denominator equals one when $|d_{i,t}|<10^{-6}$ and otherwise equals $d_{i,t}$. Ridge, random forest and histogram gradient boosting are compared using five-fold GroupKFold over the 407 calibration wells. The other 50 wells have no eligible calibration windows but remain in the 457-well test cohort. Mean per-well out-of-fold blended RMSE selects the gate; pooled out-of-fold RMSE and then candidate name resolve exact ties using calibration data only. Histogram gradient boosting is selected in all five seeds and refitted on all 6{,}562 calibration observations. The safeguard, stabilization term and fitted regressor mean that this is not an exact algebraically equivalent optimum of forecast SSE. Both the calibration targets and predicted weights are clipped to $[0,1]$. Gamma is a mixing coefficient, not a probability that transfer is beneficial.

A simpler \textit{Site-Hard Expert Selector} makes a discrete site-level choice between the same frozen No TL and Always TL parents. Its default is Always TL; it switches to No TL when calibration RMSE for Always TL exceeds No TL by more than $0.02$ and training overlap is at least 72 months. A well without a calibration policy entry uses No TL. The soft gate gives a continuous transfer-weight signal, while the hard selector is a secondary fixed routing comparison.

\subsection{Baselines and Comparison Models}

The corrected benchmark contains the \textit{Deep Context LSTM (No TL)} and \textit{Deep Context LSTM (Always TL)} parents, \textit{Global Convex Blend} and \textit{CASL}, with \textit{Site-Hard} as a secondary selector. Random Forest and the baseline LSTM family were not rerun under this protocol and supply no results for the corrected comparison.

Global Convex Blend is a non-contextual attribution comparator using the same frozen parents, with one global scalar $\alpha$ per seed. For calibration observation $j$, let $d_j=\hat{y}_j^{\mathrm{AlwaysTL}}-\hat{y}_j^{\mathrm{NoTL}}$. Its prediction and fitted weight are
\[
\hat{y}_j^{\mathrm{GCB}}=(1-\alpha)\hat{y}_j^{\mathrm{NoTL}}+\alpha\hat{y}_j^{\mathrm{AlwaysTL}},
\]
\[
\alpha=\operatorname{clip}_{[0,1]}\left(
\frac{\sum_{j\in\mathrm{cal}}d_j(y_j-\hat{y}_j^{\mathrm{NoTL}})}
{\sum_{j\in\mathrm{cal}}d_j^2}\right).
\]
This closed-form pooled calibration-SSE optimum is clipped to $[0,1]$ and frozen before test evaluation. It uses no site, state or similarity features and no validation/test tuning. The comparator is computed from saved predictions without additional parent training and is not a proposed novel model. Its calibration objective differs from CASL's weighted targets and mean-well gate-selection criterion, so the comparison does not isolate a causal contribution of contextual inputs.

\section{Experimental Setup}
\label{sec:exp_setup}

\subsection{Data Sources}

The target domain consists of Dutch BRO groundwater monitoring wells processed into a source-compatible Netherlands target panel. The panel contains all eligible Dutch BRO wells under the 1996--2016 modelling constraints. The archived candidate universe contains 496 wells. After corrected supervised $H{=}3$ construction and input-history eligibility checks, the model-window dataset retains 457 wells and 72{,}801 eligible model windows, including 13{,}718 test windows. The filtering stages leading to the final target panel are summarised in Appendix~\ref{sec:apx:methodology_details}, Table~\ref{tab:nl_data_funnel}. Before enforcing consecutive input history, this retained cohort contains 91{,}011 candidate endpoints with exact-horizon responses, as distinguished from final model windows in Table~\ref{tab:training_config}. The groundwater monitoring and groundwater-use data are obtained from the Dutch \citet{broNationalKeyRegistrySubsurface}, accessed through the DINOloket portal~\cite{dinoloketSubsurfaceData}. The pumping feature (\texttt{pumping\_\allowbreak m3\_\allowbreak month\_\allowbreak log1p}) is an approximate anthropogenic proxy rather than a measured per-well abstraction: it sums licensed groundwater-use volumes registered within a $5$\,km buffer of each well (absent registrations set to zero), divides annual volume by 12, applies facility validity dates and is log-transformed, so it should be read as a coarse indicator of local abstraction pressure.

The source domain consists of 512 gridded monthly hydroclimatic anomaly cells and 122{,}880 source rows; these gridded TSMP-G2A, COSMO-REA6, and GLEAM anomaly products follow the J\"ulich DATA pan-European groundwater monitoring data record of Ma et al.~\cite{maAdvancingAIbasedPanEuropeanDataset2021,maUsingLongShortTerm2021,maIndirectApproachBased2021,maAdvancingAIbasedPanEuropean2022}. These data provide regional context for source-domain pretraining but do not represent the same measurement process as local Netherlands target wells. This source--target mismatch may contribute to heterogeneous transfer outcomes across wells. The corrected forecasting task is exactly three-calendar-month-ahead groundwater anomaly prediction.

\subsection{Data Processing}

The corrected data processing pipeline is fixed across the five seeds. Groundwater observations are aggregated to monthly level, transformed into anomalies, aligned with hydroclimatic source data, and converted into supervised forecasting tables with lagged predictors and future targets. All predictor inputs --- the shared signals (\texttt{tsmp\_wtda}, precipitation anomaly, soil-moisture anomaly) and their calendar lags, the seasonal encodings, and the local groundwater (\texttt{wtda}) history --- have calendar dates at or before the forecast origin. For target well $i$ at time $t$, the observed groundwater level anomaly relative to NAP is defined as a standardised calendar-month anomaly,
\[
\mathrm{wtda}_{i,t} =
\frac{\ell_{i,t} - \mu^{\text{train}}_{i,m(t)}}
{\sigma^{\text{train}}_{i,m(t)}} ,
\]
where $\ell_{i,t}$ is observed water level relative to NAP, $m(t)$ is the calendar month, and $\mu^{\text{train}}_{i,m}$ and $\sigma^{\text{train}}_{i,m}$ are the site-specific training-period mean and standard deviation for calendar month $m$. Climatology uses observations through July 2008 from the original 754-well monthly panel; cohort filtering does not refit this fallback pool. Calendar-month standard deviations require three observations; zero or insufficient estimates are backed off to the site-level standard deviation (requiring 12 observations), then the global training standard deviation, and floored at a small minimum ($0.05$\,m). Missing means similarly fall back to site/global training means. Eligible endpoints require complete observed inputs at every month $t-8,\ldots,t$ and an observed response at exactly $t+3$. Missing groundwater months are neither interpolated nor replaced by later available records. Feature-history rows are retained even if they lack their own future label. Existing well eligibility requires at least 24 observed months, eight valid training windows and one valid test window. The final model inputs are complete and need no later imputation step. Monthly aggregation makes the signal more stable but smooths short-term extremes; the Netherlands target panel is treated as a reproducible empirical basis rather than a perfect representation of all possible groundwater conditions.

Shared-feature scaling is fitted on all 69{,}632 source training windows before filtering and used by both parents; context and response scaling use corrected target training windows only. Transforms are applied unchanged to validation, calibration, and test periods. No TL therefore means no source encoder pretraining, not absence of source covariates or the common source-fitted transform. The original chronological boundaries remain frozen rather than being recomputed from the nominal 60/15/10/15 ratios; calibration is reserved for selector fitting. Membership uses the exact target month: train 1996-10--2008-07, validation 2008-08--2011-06, calibration 2011-07--2013-05, and test 2013-06--2016-06, with 41{,}551, 10{,}970, 6{,}562 and 13{,}718 eligible windows, respectively. These boundaries are shared calendar ranges across wells rather than well-specific split dates. No forecast-origin embargo is imposed: the earliest validation, calibration and test origins are May 2008, April 2011 and March 2013, respectively. The inherited source training labels extend through September 2008 and source validation labels through September 2011. This is fixed target-date evaluation, not strict rolling deployment validation.

\subsection{Training Configuration}

The main context-aware parents share the same backbone family, so the intentional difference between them is encoder pretraining. Parent models use dropout 0.2, AdamW, weight decay 0.01 and gradient clipping at norm one. Source pretraining uses learning rate $10^{-3}$, at most 12 epochs, patience 3 and batch size 256; target training uses learning rate $7 \cdot 10^{-4}$, at most 28 epochs, patience 6 and batch size 128. Only the pretrained shared encoder is copied into a fresh target model, and all target parameters are fine-tuned. Best checkpoints use unweighted mean validation-batch MSE, with improvement exceeding $10^{-5}$; calibration and test labels do not select them. The selector is trained only after both parents are frozen; gate fitting updates neither parent. The selected histogram-gradient gate uses maximum depth five, up to 500 iterations, learning rate 0.03 and minimum leaf size 20. Table~\ref{tab:training_config} summarises the forecasting setup, sample sizes and selected model settings used in the experiments.

\begin{table}[!t]
\centering
\caption{Main training configuration used in the corrected experiments. Target splits retain the fixed response-date boundaries; the selected CASL gate is histogram gradient boosting (HGB).}
\label{tab:training_config}
\small
\begin{tabular}{@{}p{0.43\linewidth}p{0.49\linewidth}@{}}
\toprule
Parameter & Value \\
\midrule
Forecast horizon & 3 exact calendar months \\
Sequence length & 9 consecutive observed months \\
Target splits (train/val/cal/test) & Frozen target-date boundaries \\
Windows (train/val/cal/test) & 41{,}551 / 10{,}970 / 6{,}562 / 13{,}718 \\
Forecast-origin embargo & None \\
Source filtering rule & \texttt{maxsim0.60} (Always-TL validation) \\
Target wells & 457 \\
Test windows & 13{,}718 \\
Target exact-H=3 endpoints & 91{,}011 \\
Source cells before filtering & 512 \\
Source cells retained for pretraining & 4 \\
Candidate source rows (512 cells) & 122{,}880 \\
Source pretraining windows (train/val) & 544 / 144 \\
Shared encoder hidden size & 64 \\
Context encoder hidden size & 28 \\
Selected gate depth / minimum leaf size & 5 / 20 \\
Selected gate learning rate / maximum iterations & 0.03 / 500 \\
\bottomrule
\end{tabular}
\end{table}

\subsection{Evaluation Strategy}

The evaluation reports RMSE, MAE, and $R^2$; RMSE is the main aggregate metric because large forecasting errors matter in groundwater prediction and RMSE is sensitive to them. Pooled metrics use all test observations, while harm and rescue rates give each well equal weight. Fixed harm means Always-TL test RMSE exceeds No-TL RMSE. Among those fixed-harmed wells, partial rescue means CASL lowers RMSE relative to Always TL, and full rescue means CASL lowers it below No TL. Persistent harm means a fixed-harmed well remains worse than No TL; residual harm includes all wells where CASL is worse than No TL, including newly harmed wells where Always TL was no worse than No TL. Comparisons are strict and rescue rates use fixed-harmed wells as denominator; the new-harm rate uses Always-TL non-harmed wells. No ties occur among the core parents/CASL runs. The same definitions apply to Global Convex Blend, including equality with No TL counting as neither strict full rescue nor persistent harm. Partial rescue is descriptive: constant convex blending toward No TL with $\alpha<1$ guarantees partial rescue of fixed-harmed wells and introduces no new harm.

The corrected robustness evidence uses the predefined seeds $[11,23,42,67,89]$, recorded before the additional runs; seed 42 is reused without retraining. Corrected inputs, eligibility, architecture, splits, features and the source filter remain fixed. Initialization, training order, dropout, parent predictions and fitted gates vary by seed; calibration-only grouped gate selection uses the same rule in every seed. All seeds are reported using means, sample standard deviations and observed ranges, with no significance tests. Complete per-seed test RMSE values and calibration-fitted GCB
coefficients are provided in Table~\ref{tab:apx_seed_stability_perseed}
(Appendix~\ref{sec:apx:robustness}).These ranges describe fixed-split seed sensitivity, not population uncertainty or temporal robustness. CASL and Site-Hard validation predictions are generated after calibration fitting and are retrospective diagnostics, not chronological model-selection evidence. Rolling or blocked temporal validation has not been performed for the corrected protocol. Detailed supporting figures are collected in Appendices~\ref{sec:apx:robustness} and~\ref{sec:apx:error_analysis}.

\subsection{Reliability}

Reliability is supported by the saved corrected inputs, separate seed directories, protocol and seed manifests, hashes, predictions and well-level outcomes, together with explicit eligibility and chronological splits and a combined pooled/per-site evaluation. The legacy GitHub link does not establish public access to the corrected release; its access details remain to be confirmed. The most consequential limitations (gridded source, monthly aggregation, small calibration split) are revisited in the Discussion.

\FloatBarrier
\section{Results}
\label{sec:results}

This section reports one corrected Netherlands analysis under the exact-calendar protocol: nine consecutive observed monthly inputs and an exact three-month target, without interpolation. All five predefined seeds ($11,23,42,67,89$) evaluate the same $457$ wells and $13{,}718$ matched test windows. The main comparison includes No TL, Always TL, Global Convex Blend (GCB) and CASL; Site-Hard remains a secondary selector. Pooled accuracy and per-well harm are separate evaluation objectives. Detailed seed-42 diagnostics illustrate one reference realization rather than a representative run.

\subsection{Main Model Comparison}
\label{subsec:main_comparison}

\begin{figure*}[!t]
\centering
\includegraphics[width=\linewidth]{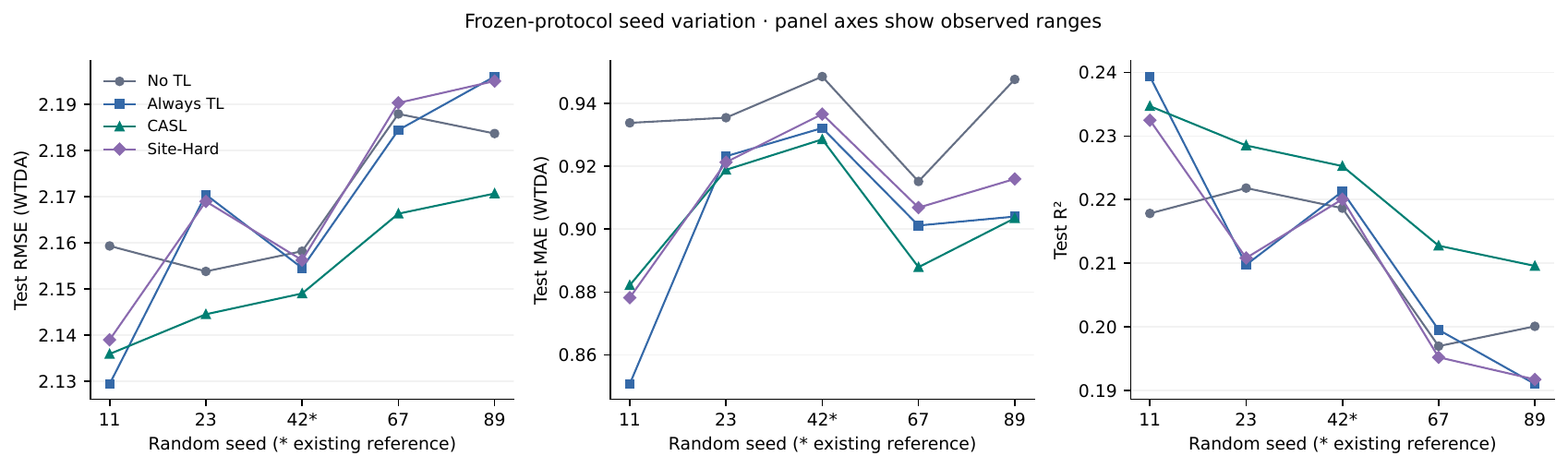}
\caption{Held-out aggregate RMSE, MAE and $R^2$ under the corrected frozen protocol for seeds 11, 23, 42, 67 and 89. Every seed evaluates the same 457 wells and 13,718 test observations. Axis ranges expose close differences; the figure does not establish large or universal superiority. CASL RMSE is lower than No TL in five seeds and lower than Always TL in four. GCB is reported in the main tables, rather than plotted here. Seed 42 is the unchanged reference; no model or seed is selected from these test scores. Observed variation describes training randomness conditional on fixed data/splits, not temporal or population uncertainty.}
\label{fig:cross_seed_aggregate}
\end{figure*}

CASL remains competitive with No TL and has slightly lower aggregate RMSE in all five seeds. Relative improvement is small: 0.424--1.084\% (mean 0.705\%). CASL has lower aggregate RMSE than Always TL in four of five seeds; its relative RMSE improvement ranges from $-0.305$\% to 1.195\%. The negative value denotes one realization in which Always TL has lower aggregate RMSE. Thus the evidence does not support universal superiority over the transfer parent or large aggregate gains (Table~\ref{tab:main_model_comparison}; Figures~\ref{fig:cross_seed_aggregate},~\ref{fig:net_harm}). Mean MAE is slightly lower for Always TL than CASL, further limiting an across-metric superiority claim. Corrected RF, baseline-LSTM and rolling-validation comparisons remain outside this evaluation.

CASL has lower aggregate RMSE than Global Convex Blend in four of five seeds, by 0.388--1.084\%. Global Convex Blend has slightly lower RMSE in seed 42, by 0.032\%; this tiny difference is not evidence of equivalence or a substantial accuracy advantage. The calibration-fitted $\alpha$ is zero in seeds 11, 67 and 89, 0.264697 in seed 23 and 0.567447 in seed 42 (mean 0.166429; range 0--0.567447). Zero weights mean that the pooled calibration optimum selects No TL for those realizations, not that transfer is generally useless. The figures display the parent/CASL comparison, with Site-Hard included in the aggregate panel; GCB is reported in Tables~\ref{tab:main_model_comparison} and~\ref{tab:cross_seed_rescue}.
\begin{table}[htbp]\centering\small
\caption{Aggregate held-out metrics across the five predefined seeds on the same 13,718 observations. Entries show mean $\pm$ sample SD; bracketed RMSE ranges are observed minima and maxima. Seed variation is conditional on the fixed data and split, not population or temporal uncertainty. Anomaly error units are dimensionless.}
\label{tab:main_model_comparison}
\resizebox{\linewidth}{!}{%
\begin{tabular}{lrrr}\toprule
Policy & RMSE (range) & MAE & $R^2$ \\\midrule
No TL & $2.16856\pm0.01593$ & $0.93611\pm0.01351$ & $0.21107\pm0.01161$ \\
 & $[2.15378,2.18791]$ & & \\
Always TL & $2.16695\pm0.02611$ & $0.90227\pm0.03155$ & $0.21219\pm0.01894$ \\
 & $[2.12938,2.19600]$ & & \\
Global Convex Blend & $2.16640\pm0.01818$ & $0.92863\pm0.01253$ & $0.21263\pm0.01323$ \\
 & $[2.14830,2.18791]$ & & \\
CASL & $2.15326\pm0.01474$ & $0.90420\pm0.01974$ & $0.22217\pm0.01066$ \\
 & $[2.13588,2.17066]$ & & \\
Site-Hard & $2.16989\pm0.02342$ & $0.91180\pm0.02164$ & $0.21007\pm0.01703$ \\
 & $[2.13895,2.19503]$ & & \\\bottomrule
\end{tabular}%
}
\end{table}

In seed 42, GCB has RMSE $2.148304$, compared with $2.148994$ for CASL, $2.154517$ for Always TL, $2.156212$ for Site-Hard and $2.158150$ for No TL. The $0.032\%$ GCB--CASL gap is tiny. The complete reference-seed comparison, including GCB, is in Appendix~\ref{sec:apx:robustness}, Table~\ref{tab:apx_transfer_risk}. RF and baseline LSTMs were not rerun and are not part of this corrected benchmark.

\subsection{Negative Transfer and Selective Recovery}
\label{subsec:negative_transfer}

For well $i$, $\Delta_i^{\mathrm{TL}}=\mathrm{RMSE}_i(\mathrm{Always\ TL})-\mathrm{RMSE}_i(\mathrm{No\ TL})$. A policy is harmed when its analogous delta is strictly positive. Partial rescue requires lower RMSE than Always TL on a fixed-harmed well; full rescue additionally requires strictly lower RMSE than No TL. Equality with No TL removes baseline-relative harm but is not full rescue.

Always TL harms a substantial but seed-variable subset: 84--237 wells, or 18.38--51.86\%. CASL partially rescues 97.53--99.50\% of these fixed-harmed wells in every seed (mean 98.48\%, SD 0.80 percentage points). This descriptive rate does not establish a benefit from contextual adaptation, because constant convex shrinkage toward No TL also guarantees partial rescue for $\alpha<1$. Full rescue is consistently present but more variable: 36--88 wells, or 23.46--43.78\% of fixed-harmed wells (mean 33.68\%, SD 9.34 percentage points; Table~\ref{tab:cross_seed_rescue}). Partial rescue denotes reduced error relative to Always TL and must not be equated with restoration of the No-TL baseline.

Across seeds, mean per-well CASL improvement over Always TL among fixed-harmed wells ranges from 0.03966 to 0.07594 anomaly RMSE units; the corresponding within-seed medians range from 0.03184 to 0.06338. These changes describe severity alongside the direction-only partial-rescue count. The harmed subset differs across seeds, so the means are not repeated measurements of an identical harmed-well group.

\begin{figure*}[!t]
\centering
\includegraphics[width=\linewidth]{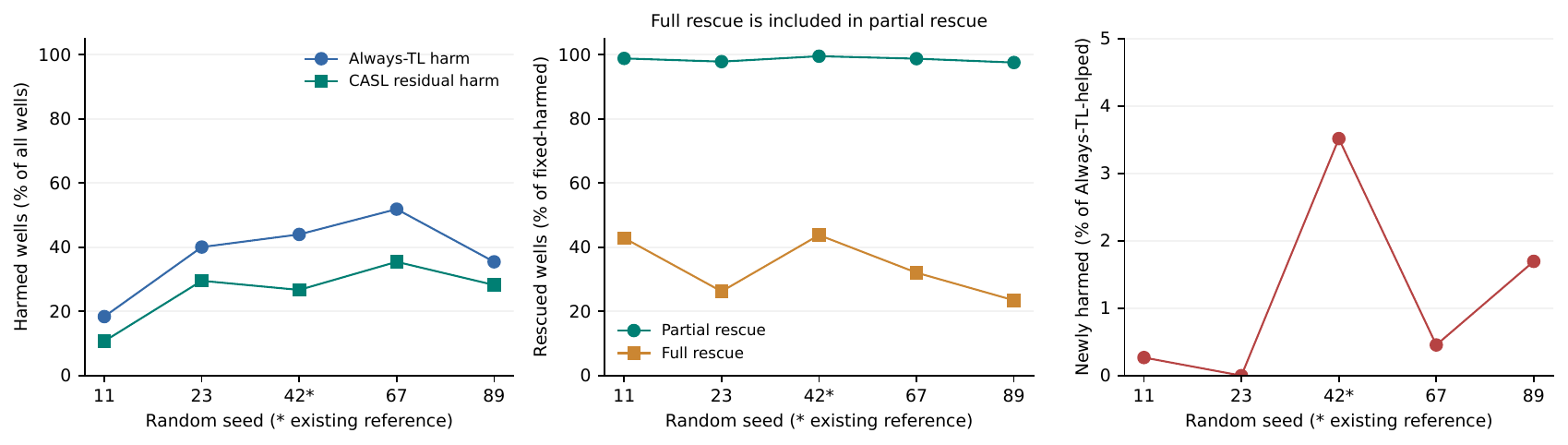}
\caption{Harm and rescue over the five predefined seeds. Always-TL harm and residual CASL harm use all 457 wells; partial/full rescue condition on the seed-specific Always-TL-harmed subset; new harm conditions on wells helped by Always TL. Full rescue is included in partial rescue and the two must not be added. Residual CASL harm includes persistent and new harm. Connected points identify seeds, not temporal change; seed 42 remains the original reference. Rates are retrospective evaluation outcomes and are not model-selection criteria. Partial rescue does not establish contextual-gate value; GCB is compared in the main tables.}
\label{fig:cross_seed_rescue}
\end{figure*}

\begin{figure*}[!t]
\centering
\includegraphics[width=\linewidth]{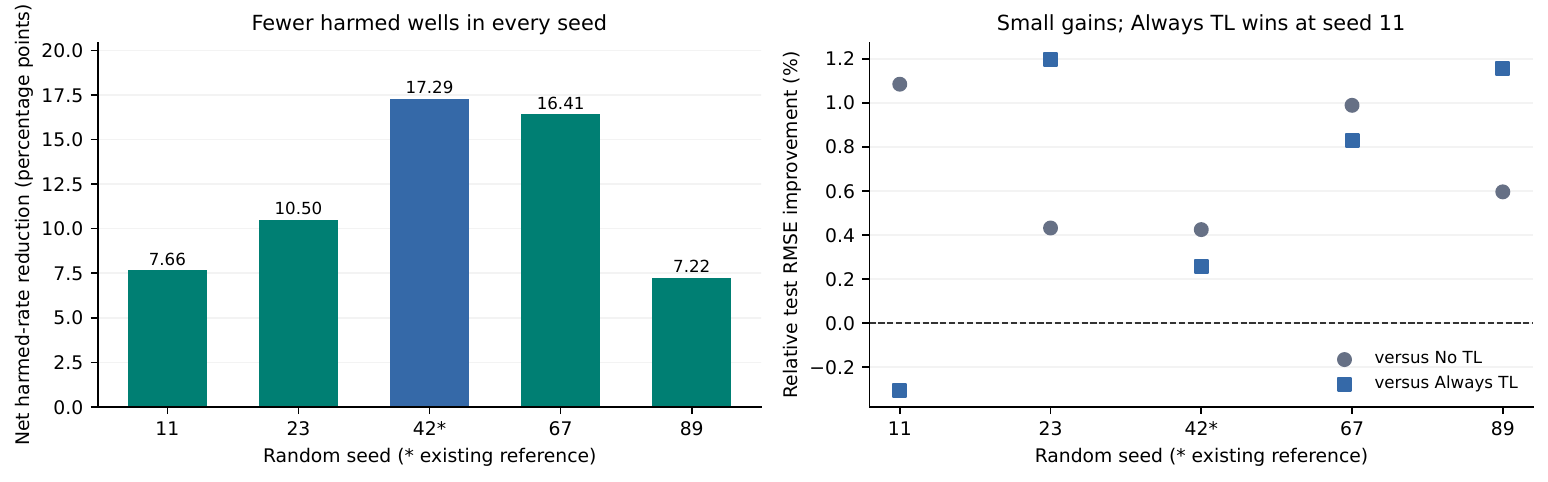}
\caption{Paired effects over the five predefined seeds under the same fixed evaluation. Left: net harmed-well rate reduction, Always-TL harm minus total CASL harm, in percentage points; the latter includes newly introduced harms. Right: relative aggregate RMSE improvements $(\mathrm{RMSE}_{\mathrm{parent}}-\mathrm{RMSE}_{\mathrm{CASL}})/\mathrm{RMSE}_{\mathrm{parent}}\times100$ versus No TL and Always TL. Positive values favor CASL. Every seed has positive net risk reduction and a small RMSE benefit versus No TL; one seed has worse aggregate RMSE than Always TL. These are paired descriptive contrasts against the parents, not evidence of stronger protection than GCB, confidence intervals or temporal robustness.}
\label{fig:net_harm}
\end{figure*}

Residual CASL harm affects 49--162 wells (10.72--35.45\%). This includes 48--161 persistent harms and zero to nine newly introduced harms per seed. New harm occurs in four seeds; its percentage among wells helped by Always TL ranges from zero to 3.52\% (mean 1.19\%). Nevertheless, the \emph{net} harmed-well rate falls by 7.22--17.29 percentage points in every seed, corresponding to 33--79 fewer harmed wells. This endpoint accounts for both rescues and new failures (Figures~\ref{fig:cross_seed_rescue},~\ref{fig:net_harm}); it is an observed matched-cohort effect, not a safety guarantee.

Global Convex Blend has lower harmed-well prevalence than CASL in four seeds; mean prevalence is 9.85\% versus 26.13\%. In seed 42, CASL harms only two fewer wells (0.44 percentage points). Global Convex Blend partially rescues all fixed-harmed wells and introduces zero new harm, consistent with the constant-blend bound. CASL records more strict full rescues in four seeds, but three comparisons are against $\alpha=0$, which exactly restores No TL and therefore counts no strict full rescue. When $\alpha>0$, full-rescue comparisons are mixed: Global Convex Blend rescues 82 versus CASL's 48 wells in seed 23 (44.81\% versus 26.23\%), whereas CASL rescues 88 versus 77 in seed 42 (43.78\% versus 38.31\%). More strict full rescue does not establish stronger protection against baseline-relative harm.
\begin{table}[htbp]\centering\small
\caption{Cross-seed harm/rescue rates. Unqualified rescue, new-harm and net-reduction rows refer to CASL; global-blend rows name the comparator explicitly. Means and sample SDs are unweighted across seeds; SD is in percentage points. Harm percentages use all 457 wells, rescue percentages use the seed-specific fixed-harmed subset, and new harm uses the Always-TL-helped subset. Full rescue is nested within partial rescue; equality with No TL is not strict full rescue. Net reduction is relative to Always TL and accounts for newly introduced harm. Partial rescue is descriptive, not evidence of contextual added value.}
\label{tab:cross_seed_rescue}
\resizebox{\linewidth}{!}{%
\begin{tabular}{p{0.42\linewidth}rrrr}\toprule
Outcome & Mean & SD & Median & Range \\\midrule
Always-TL harm (\%) & 37.94 & 12.48 & 40.04 & 18.38--51.86 \\
Partial rescue (\%) & 98.48 & 0.80 & 98.73 & 97.53--99.50 \\
Global partial rescue (\%) & 100.00 & 0.00 & 100.00 & 100.00--100.00 \\
Full rescue (\%) & 33.68 & 9.34 & 32.07 & 23.46--43.78 \\
Global full rescue (\%) & 16.62 & 22.88 & 0.00 & 0.00--44.81 \\
Residual CASL harm (\%) & 26.13 & 9.23 & 28.23 & 10.72--35.45 \\
Residual global-blend harm (\%) & 9.85 & 13.60 & 0.00 & 0.00--27.13 \\
New harm among helped (\%) & 1.19 & 1.46 & 0.45 & 0.00--3.52 \\
Global new harm among helped (\%) & 0.00 & 0.00 & 0.00 & 0.00--0.00 \\
Net harmed-rate reduction (pp) & 11.82 & 4.77 & 10.50 & 7.22--17.29 \\\bottomrule
\end{tabular}%
}
\end{table}

\subsection{Robustness}
\label{subsec:robustness}

All predefined runs completed successfully, and seed 42 was retained without retraining. Data, features, source filter, architecture and split definitions remain frozen; initialization, optimization and fitted gate realization vary by seed. Means, sample standard deviations and observed ranges describe training randomness on this fixed split, not population uncertainty, independent spatial replication or temporal deployment robustness. No corrected-protocol significance claim is made, and rolling/blocked temporal validation has not been completed. Per-seed RMSE and calibration $\alpha$, together with reference-seed harm/rescue outcomes, are reported in Appendix~\ref{sec:apx:robustness}.

\begin{figure*}[!t]
\centering
\includegraphics[width=\linewidth]{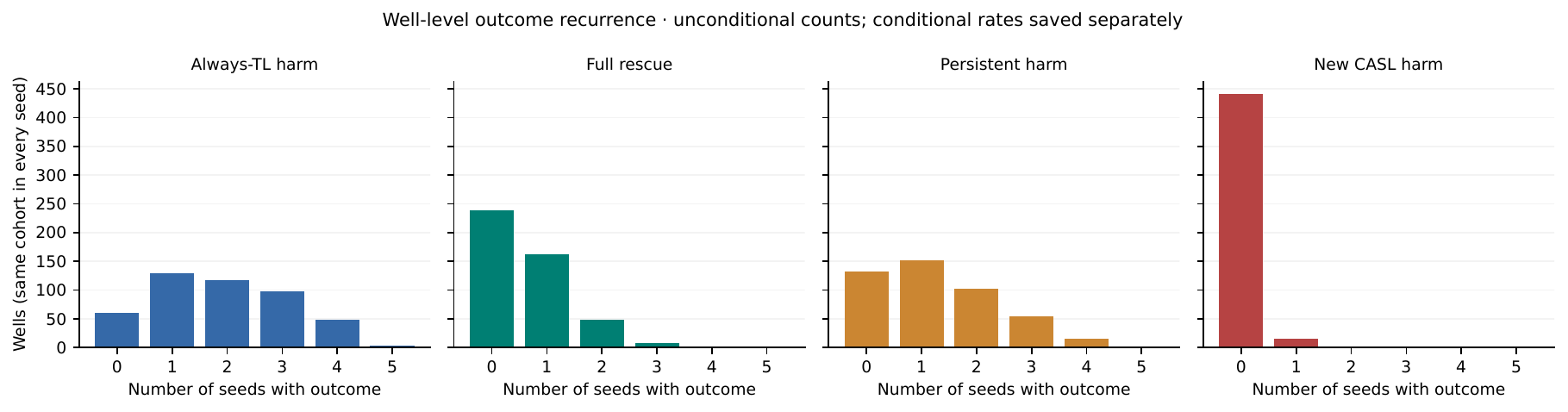}
\caption{Distribution of the number of seeds in which a well has each outcome. All 457 wells are retained. Full rescue and persistent harm require Always-TL harm in the same seed; their per-well conditional eligibility counts are retained in the outcome-frequency CSV. Only three wells are harmed by Always TL in all five seeds; no well is fully rescued or persistently harmed in all five. Sixteen distinct wells are newly harmed once each, with no repeated new-harm identity. Recurrence of the CASL-versus-Always-TL outcomes does not imply stable individual identities or superior protection over GCB.}
\label{fig:well_recurrence}
\end{figure*}

Negative transfer is a recurring cohort-level phenomenon, but individual affected-well identities vary. Always TL harms 267 wells in at least two seeds; only three wells are harmed in all five. Full rescue recurs at 57 wells, and persistent harm recurs at 172 wells, in at least two seeds (Figure~\ref{fig:well_recurrence}). Neither status appears at any well in all five seeds. Recurrence counts must be read with their eligibility denominators: full rescue or persistent harm is possible only in seeds where Always TL harms that well. Well identifiers and complete outcome matrices accompany the saved corrected analysis.

Sixteen distinct wells are newly harmed by CASL at least once, and each shows that status in exactly one seed. This does not establish a fixed class of newly vulnerable wells or a general failure mechanism. Well-by-seed matrices and per-well frequency tables retain all outcomes; individual identifiers are descriptive bookkeeping, not evidence of hydrogeological causes.

Test support is identical across seeds: median 33 observations per well, interquartile range 24--37 and range 1--37. Five wells have fewer than ten observations and remain included. Seed 42 has the highest partial/full rescue and net harm-rate reduction, but the smallest relative CASL RMSE improvement over No TL. It is one realization rather than a representative run; its detailed support, severity and failure-case diagnostics are appendix illustrations.

\subsection{Source Filtering Sensitivity}
\label{subsec:source_filter}

The source filter \texttt{maxsim0.60} is selected once using seed-42 Always-TL validation RMSE only, then frozen before calibration gate fitting and for all five seeds. Neither CASL validation RMSE nor calibration/test results enter this choice. The four candidates retain $512$, $230$, $69$ and $4$ cells for no filter, \texttt{maxsim0.40}, \texttt{maxsim0.50} and \texttt{maxsim0.60}, with validation RMSE $1.555131$, $1.560149$, $1.547691$ and $1.546540$, respectively. The two leading candidates are numerically near-equivalent: the gap between \texttt{maxsim0.50} and \texttt{maxsim0.60} is only $0.001150$ RMSE (about $0.074\%$). This operational choice does not establish a scientifically optimal threshold. The complete corrected candidate table is in Appendix~\ref{sec:apx:methodology_details}, Table~\ref{tab:apx_source_filter}.

\subsection{Selector Interpretation}
\label{subsec:selector_interp}

CASL fits a contextual blending weight on calibration labels after both parent models are frozen. Its input context and nonconstant weights do not by themselves establish added value over non-contextual blending.

Within each seed's fixed-harmed subset, recent anomaly magnitude, recent target-series endpoint change and retrospective standardized train--test mean shift have positive Spearman associations with CASL-minus-No-TL RMSE in all five seeds. Their ranges are respectively 0.042--0.399, 0.038--0.287 and 0.049--0.408: direction repeats, but strength varies and is sometimes small. Best source-similarity correlations remain weak and sign-variable: $-0.080$ to 0.054 across all 512 cells and $-0.056$ to 0.072 within the frozen four-cell pool.

Related group comparisons appear in Appendix Figure~\ref{fig:apx_exploratory_patterns}; those Cliff's-delta contrasts are distinct from the Spearman associations above. Recent-state and retrospective-shift descriptors show repeated positive associations within fixed-harmed subsets, whereas source-similarity associations are weak and sign-variable. These comparisons do not establish relative predictive importance, causal mechanisms or prospective failure detection.

\subsection{Gate Behavior and Functional Sensitivity}
\label{subsec:selector_sensitivity}

The unchanged calibration-only procedure chooses \texttt{hgb\_env} in every seed. Test mean $\gamma$ ranges from 0.425 to 0.605 and its within-seed standard deviation from 0.177 to 0.246. The proportion favoring No TL ranges from 27.33\% to 65.45\%; the complementary Always-TL proportion ranges from 34.55\% to 72.67\%. None of the gates is constant, all zero/all one or confined to one side of the parent-preference boundary. No seed therefore shows the checked forms of gate collapse. Intermediate weights remain mixing coefficients, not evidence of per-well safety or calibrated event probabilities. Reference-seed distributions and inference-only group permutations are provided in Appendix~\ref{sec:apx:error_analysis} as reference illustrations rather than multi-seed training ablations.

The corrected seed-42 inference-only diagnostic jointly permutes nine disjoint groups covering the gate's 47 inputs, without refitting. Recent groundwater state and recent hydroclimatic state produce the largest observed changes in $\gamma$ ($0.1301$ and $0.1240$, equal-well-weight mean absolute change); similarity context follows ($0.0935$). Group size, dependence and unrealistic permuted combinations affect these values. This is functional sensitivity of the fitted gate, not a training ablation, conditional feature importance, causal explanation or evidence of improved accuracy. Appendix~\ref{sec:apx:error_analysis} contains the saved diagnostic figure.

\subsection{Error Analysis}
\label{subsec:error_analysis}

On the $2{,}759$ seed-42 observations with $|wtda|>1.5$, the four original deep policies have similar large errors (Appendix~\ref{sec:apx:error_analysis}, Table~\ref{tab:apx_extreme_anomaly}). These are retrospective anomaly-magnitude groups, not validated drought events. No GCB anomaly-stratum metric is reported in the saved analysis. Test support has median $33$ observations per well, interquartile range $24$--$37$ and range $1$--$37$; five wells have fewer than ten observations and remain included. All nine seed-42 newly harmed wells have at least ten test observations, so low support does not explain all failures. Origin-state, test-shift and registry-context diagnostics remain descriptive; none establishes operational drought-warning validity or prospective failure detection.

Across five seeds, fixed Always TL harms $18.38$--$51.86\%$
of wells. CASL partially rescues $97.53$--$99.50\%$ and fully
rescues $23.46$--$43.78\%$ of those wells, although persistent
and newly induced harm remain. GCB has lower harmed-well
prevalence in four seeds, whereas CASL has slightly lower
pooled RMSE in four. Neither policy dominates both measures.
Source--target similarity shows weak and sign-variable
associations with residual CASL harm across seeds.
\FloatBarrier
\section{Discussion}
\label{sec:discussion}

\subsection{Negative Transfer and the Accuracy--Risk Trade-off}

The empirical contribution of selective transfer in this study is not a large accuracy jump over fixed transfer. Across five seeds on the corrected Netherlands target dataset, Context-Aware Selective Learning has lower pooled RMSE than No TL in all five seeds and Always TL in four, but its improvement over No TL is only $0.424$--$1.084\%$. The more important result sits at the site level. The fixed-transfer Always TL parent harms $18.38$--$51.86\%$ of the $457$ target wells; Context-Aware Selective Learning partially rescues $97.53$--$99.50\%$ of those harmed wells and fully rescues $23.46$--$43.78\%$. These are descriptive rates: partial rescue can arise from convex blending towards No TL without contextual adaptation. Global Convex Blend protects more wells from harm in four of five seeds, while CASL has slightly lower pooled RMSE than GCB in four seeds. Neither dominates, so the result is an accuracy--risk trade-off rather than stronger harm control attributable to context.Supporting seed-42 diagnostics illustrate aggregate performance
(Figure~\ref{fig:apx_seed42_aggregate}), harm and rescue rates
(Figure~\ref{fig:apx_seed42_rescue}), outcome categories
(Figure~\ref{fig:apx_seed42_taxonomy}), and error reductions among
fixed-harmed wells (Figure~\ref{fig:apx_seed42_mitigation})
in Appendix~\ref{sec:apx:robustness}. \citet{profetaPredictingGroundwaterLevel2025a} reported a clearer aggregate transfer-learning gain in a Dutch regional setup; on the broader Netherlands panel, the aggregate gain here is smaller. Earlier groundwater transfer-learning work also rarely audits site-level transfer outcomes, so this study is best framed as an evaluation design that takes negative transfer seriously and compares learned contextual blending with ordinary convex shrinkage.

\subsection{Why the Aggregate Gain is Small but Still Meaningful}

The closeness of the parent and blending policies on pooled RMSE does not make their well-level outcomes interchangeable. Pooled RMSE alone hides well-level transfer harm, which is why the study reports pooled metrics together with per-well diagnostics. The corrected five-seed results show a small CASL accuracy improvement over No TL, but not universal superiority over Always TL or GCB. High partial rescue does not establish contextual added value: a constant convex weight below one structurally improves on Always TL at wells harmed by that parent. GCB's zero weight in seeds 11, 67 and 89 restores No TL exactly, removing transfer harm but producing no strict full rescues. Among the two seeds with positive GCB weights, full-rescue comparisons are mixed.

\subsection{Metric-Dependent Model Preference}

A consistent pattern across the comparison is that the best model depends on the loss function. Always TL has slightly lower mean pooled MAE than CASL across seeds ($0.90227$ versus $0.90420$), while CASL more often has slightly lower pooled RMSE. GCB has lower mean harmed-well prevalence than CASL ($9.85\%$ versus $26.13\%$), despite CASL having lower pooled RMSE in four seeds. These preferences are not contradictory: pooled prediction error and well-level harm answer different questions. The practical implication is that model comparison should follow the loss or risk measure of interest rather than assuming one model dominates under every metric. The corrected seed-42 per-policy comparison is tabulated in Appendix~\ref{sec:apx:robustness} (Table~\ref{tab:apx_transfer_risk}).

\subsection{Source--Target Mismatch and Source Filtering}

The source is gridded hydroclimatic model output rather than observed groundwater wells. This source--target mismatch may contribute to heterogeneous transfer outcomes across wells, but the descriptive comparisons do not establish that mechanism. The source filter is validation-selected for the Netherlands target dataset using only corrected Always TL validation RMSE, with calibration reserved for policy fitting. At \texttt{maxsim0.40} the corrected filter retains $230$ of $512$ source cells; at \texttt{maxsim0.50} $69$ survive; and at \texttt{maxsim0.60} $4$ survive. The selected \texttt{maxsim0.60} filter is frozen before calibration and remains unchanged across seeds. Its validation advantage over \texttt{maxsim0.50} is only $0.001150$ RMSE, so the leading candidates are near-equivalent numerically rather than evidence of a scientifically optimal threshold. CASL validation predictions are post-calibration diagnostics and do not select the filter. The choice is specific to this frozen comparison and does not establish a generally optimal source-similarity threshold.

The corrected descriptive comparisons do not establish which selector input is most important. Recent-state and retrospective train--test shift descriptors show repeated associations with residual CASL harm across seeds, while source--target similarity associations are weak and sign-variable. Appendix~\ref{sec:apx:error_analysis} provides the exploratory
cross-seed group contrasts
(Figure~\ref{fig:apx_exploratory_patterns}) and the illustrative
seed-42 descriptor distributions
(Figure~\ref{fig:apx_success_failure}). Similarity remains a screening input and one selector covariate, but it is not a standalone explanation of transfer success. These comparisons do not establish relative predictive importance or prospective failure detection.The inference-only gate sensitivity analysis is provided in
Figure~\ref{fig:apx_gate_sensitivity}; it describes changes in
blending weights under grouped input permutations rather than
a training ablation. They should be interpreted as descriptive associations rather than as evidence of a physical causal mechanism. These diagnostics do not establish a relationship between similarity and fixed-transfer gains.

\subsection{Operational Implications}

For a deployed forecasting system, average RMSE is rarely the only criterion that matters: fixed transfer improves many wells but harms a substantial subset, and that harm is invisible in a single pooled metric. Context-Aware Selective Learning makes this distinction visible, but the comparison with GCB shows that contextual adaptation does not give stronger well-level protection. Its blending weight is a continuous mixing coefficient indicating how much each frozen parent contributes, and the Site-Hard variant gives a simpler discrete version with an explicit no-transfer fallback. Reporting this weight alongside the forecast would let a downstream user audit which parent contributes more; it is not a probability of transfer safety or a physical-causality model. The illustrative seed-42 distribution of blending weights is
shown in Figure~\ref{fig:apx_gate_behavior}
(Appendix~\ref{sec:apx:error_analysis}).

The operational reach is nonetheless limited. No drought-event evaluation has been performed under the corrected protocol. Table~\ref{tab:apx_extreme_anomaly}
(Appendix~\ref{sec:apx:error_analysis}) reports retrospective
errors for observations with $|wtda|>1.5$; this magnitude-based
subset does not constitute a drought-event evaluation. This study should therefore be read as a forecasting and transfer-policy comparison, not as a complete operational drought early-warning system; reaching the latter would require explicit event definitions, calibrated detection thresholds, and evaluated recall, precision, false alarms and lead time. The fixed target-date splits have no forecast-origin embargo, so fitted models cannot be assumed to have been available at every held-out origin. The results support fixed-split robustness rather than strict rolling-origin deployment validity. Within this evaluated cohort, safer transfer relative to No TL and the smallest pooled prediction error are distinct objectives; neither GCB nor CASL dominates both.

\subsection{Limitations and Future Work}

The findings apply primarily to comparably monitored groundwater
wells, given the eligible Netherlands panel, gridded source data
and monthly resolution. Evaluation across five seeds captures
training variability on the same wells and fixed temporal split;
it does not establish generalisation to other regions or periods.
Unequal test support across wells and seed-sensitive outcomes
also warrant caution when interpreting individual-well results.
Furthermore, CASL and GCB use different calibration objectives,
so their comparison does not isolate the contribution of
contextual inputs.

Future work should prioritise rolling-origin evaluation and
selector ablations under matched calibration objectives.
Additional baseline comparisons and attention-based or hybrid
spatiotemporal models could assess whether the observed
accuracy--risk trade-off persists across model architectures.
Richer physical and anthropogenic features, longer observational
records and uncertainty estimation for transfer weights could
further support selective transfer. Moving calibration windows
could also be evaluated for their potential to improve selector
stability.
\FloatBarrier
\section{Conclusion}
\label{sec:conclusion}

This study asked whether transfer learning is conditionally useful for multi-well groundwater anomaly forecasting in the Netherlands, and whether a context-aware selective strategy can reduce negative transfer while matching or improving no-transfer and fixed-transfer baselines under the corrected protocol. The answer is conditional rather than categorical: transfer helps many wells, but the fixed-transfer Deep Context LSTM (Always TL) parent harms a substantial subset ($18.38$--$51.86\%$ of the $457$ target wells across five seeds), so transfer is best treated as a well-level decision rather than a fixed rule. Context-Aware Selective Learning gives a small accuracy gain over No TL ($0.424$--$1.084\%$ lower RMSE), partially rescues $97.53$--$99.50\%$ of fixed-harmed wells and fully rescues $23.46$--$43.78\%$. However, GCB protects more wells from harm in four seeds, while CASL has slightly lower pooled RMSE than GCB in four: neither dominates. High partial rescue alone is not evidence for contextual adaptation because convex blending towards No TL can provide it structurally. Source--target similarity was investigated descriptively, with weak and sign-variable associations with residual CASL harm. The main hypothesis receives partial support: CASL reduces well-level harm relative to Always TL while remaining competitive in pooled accuracy, but the Global Convex Blend comparison does not support stronger harm control attributable to contextual adaptation.

These conclusions are bounded by the source-compatible panel, the gridded source domain, and monthly aggregation (Section~\ref{sec:discussion}), and the system is best understood as a forecasting and transfer-policy comparison rather than a complete drought early-warning system. The corrected five-seed check covers aggregate and per-well outcomes on the same fixed split; it does not establish strict rolling-origin deployment validity. CASL remains competitive, but the aggregate gains are small and the main result is an accuracy--risk trade-off. Recent-state and retrospective-shift descriptors show repeated associations, while similarity associations with residual CASL harm are weak and sign-variable; these are descriptive rather than causal or prospective predictors. Promising next steps include stronger backbones, richer physical and anthropogenic context, and extreme-event validation, which together could move the framework from a controlled methodological comparison toward more operational decision support. Overall, transfer is not universally beneficial: CASL remains competitive and more often gives slightly lower pooled RMSE, while GCB is more conservative in harmed-well protection. Neither policy dominates; the main contribution is understanding the accuracy--risk trade-off and heterogeneous transfer outcomes across wells.

\FloatBarrier

% References
\clearpage
\bibliographystyle{cas-model2-names}
\bibliography{bibliographies/references}

% Appendices
\clearpage
\appendix
\section*{Appendices}
\suppressfloats[t]

% You can choose whether you prefer a single or double column appendix.
% Whatever you choose, you will need to stick to it throughout the appendix.
% For double column style, comment the next line.

\section{Additional Methodology Details}
\label{sec:apx:methodology_details}

This appendix collects supporting methodology details for Sections~\ref{sec:methodology} and~\ref{sec:exp_setup}, including the target-data filtering funnel, main training configuration, corrected source-filter comparison, and a compact CASL pipeline schematic. GCB and Site-Hard use the same frozen parents; their calibration rules are specified in the main Methodology.

\begin{table}[h]
\centering
\caption{Target-data filtering funnel for the Netherlands modelling panel. The original download and anomaly-panel stages are unchanged; final eligibility and test counts use the corrected calendar protocol.}
\label{tab:nl_data_funnel}
\small
\begin{tabular}{@{}lr@{}}
\toprule
Stage & Count \\
\midrule
National GLD series with CSV URLs & 112{,}627 \\
After temporal filter (first date $\le$ 2005-01-01, last date $\ge$ 2014-01-01, observations $\ge$ 500) & 2{,}102 \\
One tube per well after deduplication & 1{,}280 \\
Successful downloads / cached series & 1{,}118 \\
Series with at least 24 monthly anomaly rows in the 1996-07--2016-06 window & 754 \\
Series in the original monthly anomaly reference pool & 754 \\
Final corrected modelling wells with sufficient train and test windows & 457 \\
Test windows & 13{,}718 \\
\bottomrule
\end{tabular}
\end{table}

\begin{table}[htbp]
\centering
\caption{Corrected source-filter selection using only seed-42 Always-TL validation RMSE. The filter is frozen before gate fitting and is not reselected by seed. Gaps use unrounded scores; the leading candidates are numerically near-equivalent.}
\label{tab:apx_source_filter}
\small
\begin{tabular}{lrrrrl}
\toprule
Candidate & Source cells & Val. RMSE & Val. MAE & Gap to best & Selected \\
\midrule
\texttt{all} & 512 & 1.555131 & 0.766569 & 0.008591 & No \\
\texttt{maxsim0.40} & 230 & 1.560149 & 0.771671 & 0.013608 & No \\
\texttt{maxsim0.50} & 69 & 1.547691 & 0.753573 & 0.001150 & No \\
\texttt{maxsim0.60} & 4 & 1.546540 & 0.750084 & 0.000000 & Yes \\
\bottomrule
\end{tabular}
\end{table}

\FloatBarrier
\section{Robustness and Sensitivity Figures}
\label{sec:apx:robustness}

The main Results report five-seed aggregate performance and harm/rescue behavior. This appendix provides complete per-seed RMSE and calibration $\alpha$, plus the corrected seed-42 diagnostics. Seed 42 is illustrative: it has the highest CASL partial/full rescue and net harmed-rate reduction, and the smallest CASL RMSE improvement relative to No TL. The saved diagnostic figures show the original parent/CASL/Site-Hard comparison and omit GCB; GCB is included in the comparison tables. No model is fitted or selected from test labels.

\begin{table}[htbp]
\centering
\caption{Corrected seed-42 aggregate and transfer-risk comparison on 13,718 observations from 457 wells, including GCB. Harm counts/rates use all 457 wells; partial/full rescue use 201 fixed-harmed wells; new harm uses 256 Always-TL-helped wells. Rates are percentages. Equality with No TL is not strict full rescue. Dashes mark reference/inapplicable entries or Site-Hard diagnostics not reported here.}
\label{tab:apx_transfer_risk}
\small
\resizebox{\linewidth}{!}{%
\begin{tabular}{lrrrrrrr}
\toprule
Policy & RMSE & MAE & $R^2$ & Harm $n$ (\%) & Partial $n$ (\%) & Full $n$ (\%) & New $n$ (\%) \\
\midrule
No TL & 2.158150 & 0.948517 & 0.218662 & --- & --- & --- & --- \\
Always TL & 2.154517 & 0.932157 & 0.221290 & 201 (43.98) & --- & --- & --- \\
Global Convex Blend & 2.148304 & 0.924237 & 0.225775 & 124 (27.13) & 201 (100.00) & 77 (38.31) & 0 (0.00) \\
CASL & 2.148994 & 0.928552 & 0.225278 & 122 (26.70) & 200 (99.50) & 88 (43.78) & 9 (3.52) \\
Site-Hard & 2.156212 & 0.936597 & 0.220065 & --- & --- & --- & --- \\
\bottomrule
\end{tabular}%
}
\end{table}

\begin{table}[htbp]
\centering
\caption{Per-seed corrected test RMSE and calibration-only GCB coefficient. Every seed uses identical wells, test windows and frozen source filter. Seed 42 is the retained reference, not a representative realization. Zero $\alpha$ reproduces No TL exactly.}
\label{tab:apx_seed_stability_perseed}
\small
\begin{tabular}{rrrrrrr}
\toprule
Seed & No TL & Always TL & GCB & CASL & Site-Hard & GCB $\alpha$ \\
\midrule
11 & 2.159294 & 2.129380 & 2.159294 & 2.135883 & 2.138950 & 0.000000 \\
23 & 2.153781 & 2.170417 & 2.152840 & 2.144486 & 2.168945 & 0.264697 \\
42 & 2.158150 & 2.154517 & 2.148304 & 2.148994 & 2.156212 & 0.567447 \\
67 & 2.187909 & 2.184416 & 2.187909 & 2.166285 & 2.190298 & 0.000000 \\
89 & 2.183663 & 2.196001 & 2.183663 & 2.170657 & 2.195027 & 0.000000 \\
\bottomrule
\end{tabular}
\end{table}

In seed 42, CASL residual harm comprises 113 persistent harms and nine newly harmed wells; GCB has 124 persistent harms and zero new harm. CASL's mean and median per-well improvement over Always TL on the 201 fixed-harmed wells are $0.04590$ and $0.03799$. The five mutually exclusive CASL outcomes are 88 full rescues, 112 partial-only rescues, one without partial rescue, 247 preserved benefits and nine new harms. These classes describe CASL versus its parents; GCB's baseline ties in zero-$\alpha$ seeds require separate equality handling.

\begin{figure}[htbp]
\centering
\includegraphics[width=0.80\linewidth]{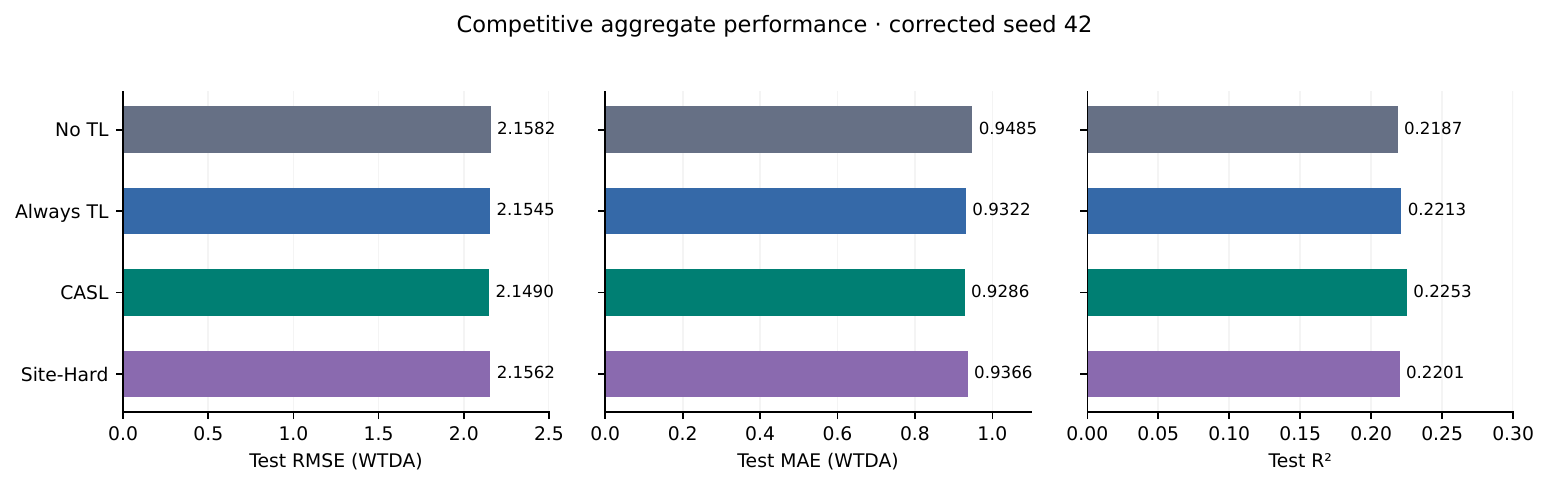}
\caption{Corrected seed-42 held-out test RMSE, MAE and $R^2$ for matched No TL and Always TL parents, their frozen CASL convex blend and secondary Site-Hard policy. All use 457 wells and 13,718 exact-H3, consecutive-nine-month test windows. Point estimates are close; this panel shows the original four-policy reference comparison only. GCB and the five-seed accuracy--risk comparison are reported in the main comparison tables. WTDA is a dimensionless standardized groundwater-level anomaly.}
\label{fig:apx_seed42_aggregate}
\end{figure}

\begin{figure}[htbp]
\centering
\includegraphics[width=0.80\linewidth]{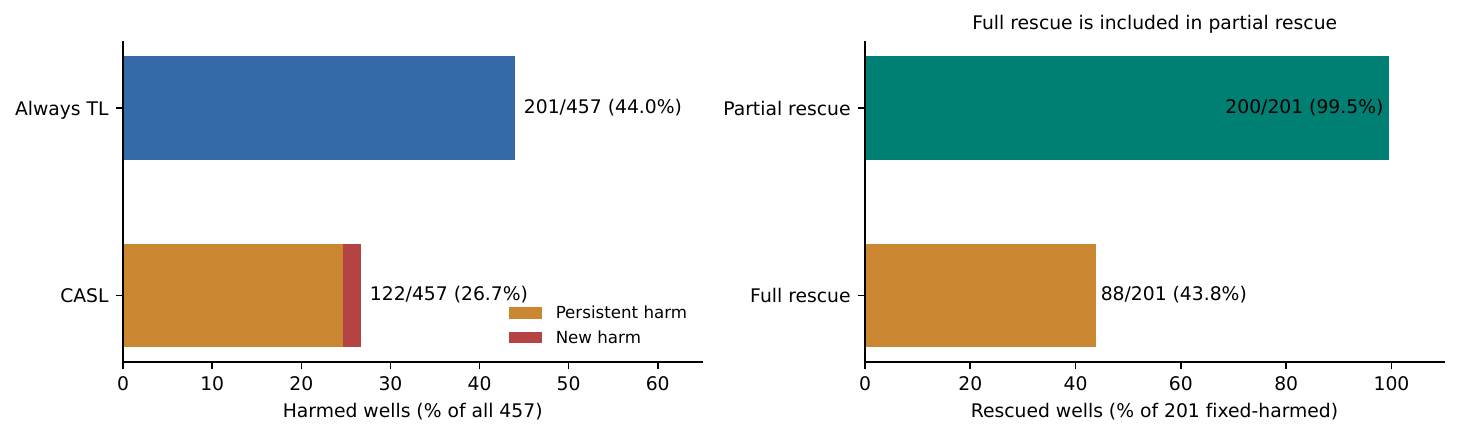}
\caption{Illustrative corrected seed-42 diagnostic. Left: strict per-well harm (policy RMSE exceeds No-TL RMSE), denominator 457 wells. CASL residual harm separates 113 persistently harmed and nine newly harmed wells. Right: partial rescue (CASL RMSE below Always-TL RMSE) and full rescue (CASL RMSE below No-TL RMSE), denominator 201 fixed-harmed wells. Full rescue is included in partial rescue; the rescue bars are not disjoint. Partial rescue is descriptive and does not establish contextual added value over global blending.}
\label{fig:apx_seed42_rescue}
\end{figure}

\begin{figure}[htbp]
\centering
\includegraphics[width=0.80\linewidth]{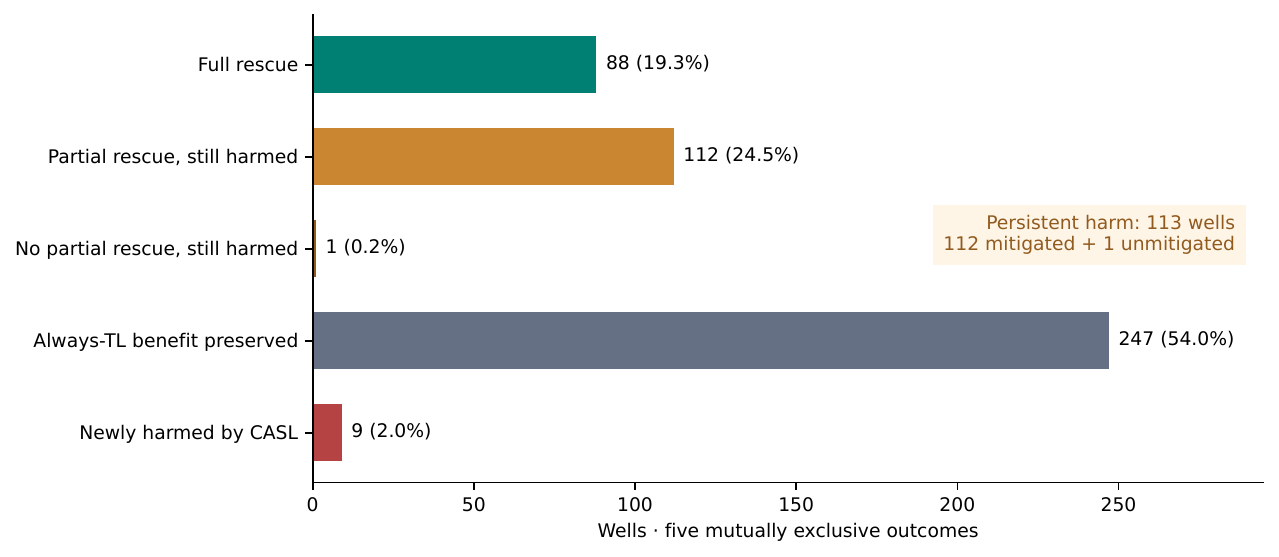}
\caption{Illustrative corrected seed-42 diagnostic. Five mutually exclusive outcomes sum to 457 wells: 88 full rescues, 112 partial-but-not-full rescues, one without partial rescue, 247 preserved Always-TL benefits and nine new harms. The 113 persistently harmed wells comprise the 112 partial-only rescues and one unmitigated case. Partial mitigation must not be equated with elimination of harm. This taxonomy applies to the three core policies in seed 42; GCB baseline ties have separate equality handling.}
\label{fig:apx_seed42_taxonomy}
\end{figure}

\begin{figure}[htbp]
\centering
\includegraphics[width=0.80\linewidth]{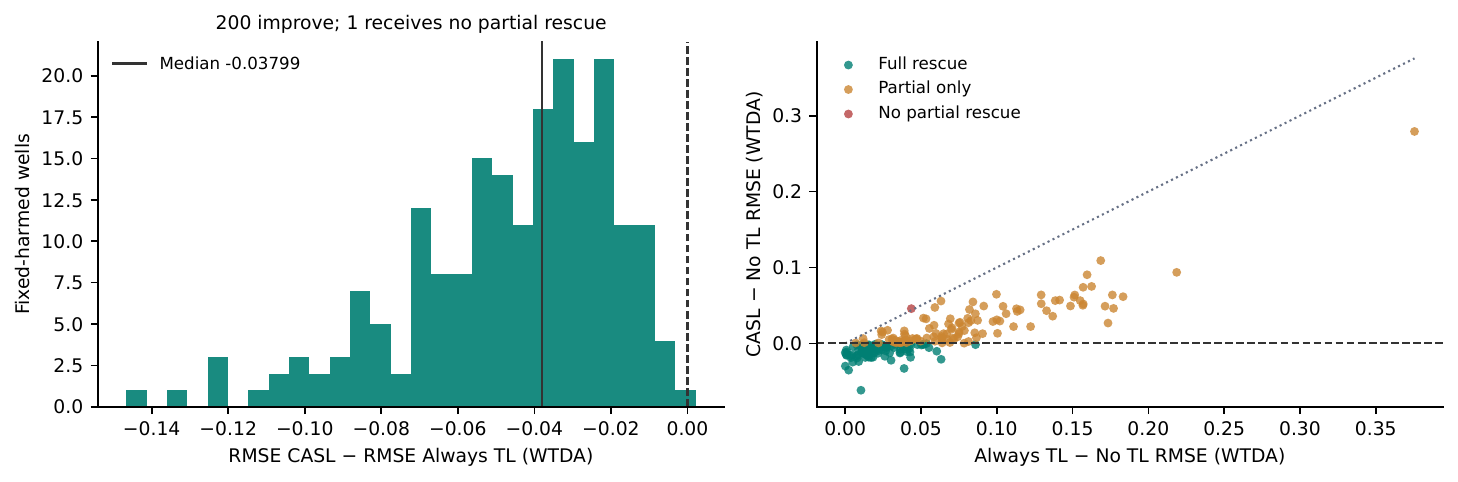}
\caption{Illustrative corrected seed-42 diagnostic. CASL versus Always-TL RMSE changes on the same 201 fixed-harmed wells, with equal well weight. Negative histogram values indicate reduced error; the median is $-0.03799$. The scatter compares excess RMSE relative to No TL before and after blending: points below zero are full rescues and points below the diagonal receive partial mitigation. These are descriptive changes relative to Always TL, not evidence that CASL provides stronger protection than GCB.}
\label{fig:apx_seed42_mitigation}
\end{figure}

\FloatBarrier
\section{Error Analysis Details}
\label{sec:apx:error_analysis}

These corrected diagnostics distinguish forecast-origin inputs from retrospective descriptors that use realized test truth. Recent anomaly magnitude is $|a_{i,t}|$; recent change is $|a_{i,t}-a_{i,t-8}|/8$, averaged over each well's evaluated forecasts. Retrospective shift is the absolute difference between test and training target means divided by the training target standard deviation. Source similarity uses training histories. Group comparisons and correlations are descriptive, not causal explanations, relative predictive-importance estimates or prospective failure detectors.

The cross-seed exploratory figure uses Cliff's-delta group contrasts, which differ from the within-fixed-harm Spearman associations in the main Results. Recent-change full-rescue-versus-persistent contrasts repeat direction in five seeds, whereas magnitude and shift contrasts repeat in four. Newly harmed groups contain one, zero, nine, one and five wells in seed order; the seed-23 contrast is undefined. The same wells and histories are reused across seeds, so these are not independent ecological replications.

\begin{figure}[htbp]
\centering
\includegraphics[width=0.80\linewidth]{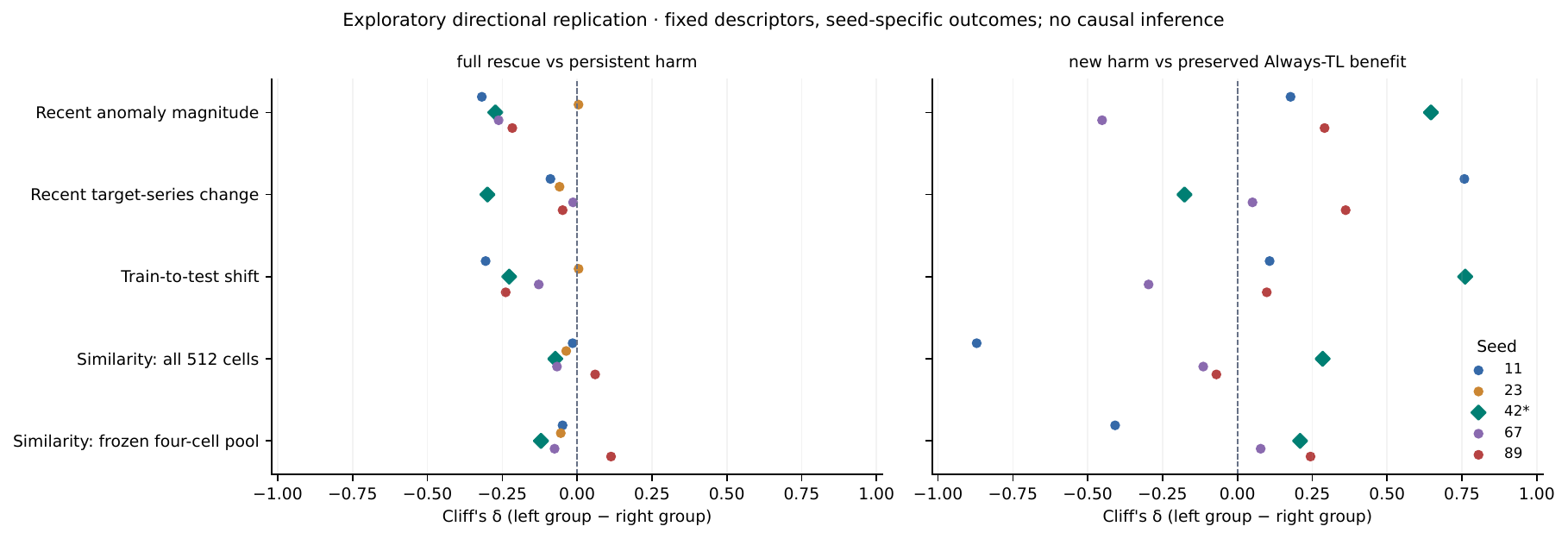}
\caption{Descriptive Cliff's delta for the same previously available well descriptors and seed-specific outcome groups. Positive values denote larger descriptor values in the left group. Train--test shift uses realized test truth; source similarity uses training histories. Seed agreement describes outcome sensitivity to training randomness on the same cohort, not independent ecological replication or causal evidence. Magnitude/shift full-rescue-versus-persistent contrasts repeat direction in four seeds; recent change repeats in five. Within-fixed-harm Spearman associations, discussed separately in the text, are a different estimand. Seed 23 has no newly harmed wells and no right-panel contrast; remaining new-harm groups contain only one to nine wells. The seed-42 new-harm profile is illustrative rather than a general mechanism. These comparisons do not establish relative predictive importance or added protection from contextual features.}
\label{fig:apx_exploratory_patterns}
\end{figure}

The seed-42 support distribution has median 33 observations per well, interquartile range 24--37 and range 1--37. Five wells have fewer than ten observations: four are fixed-harmed, all four receive partial rescue and one receives full rescue; three remain harmed by CASL. All nine newly harmed wells have 15--37 observations. No descriptive support stratum changes model eligibility.

\begin{table}[htbp]
\centering
\caption{Corrected seed-42 retrospective errors on the 2,759 test observations with $|wtda|>1.5$. These anomaly-magnitude groups are not validated drought events. GCB stratum errors were not computed in the saved analysis and remain unreported. No event-detection metrics are implied.}
\label{tab:apx_extreme_anomaly}
\small
\begin{tabular}{lrr}
\toprule
Policy & RMSE & MAE \\
\midrule
No TL & 4.488063 & 1.949304 \\
Always TL & 4.477892 & 1.906858 \\
Global Convex Blend & --- & --- \\
CASL & 4.476309 & 1.916308 \\
Site-Hard & 4.483016 & 1.909162 \\
\bottomrule
\end{tabular}
\end{table}

Seed-42 full-rescue versus persistent-harm median origin anomaly magnitude is $0.6733$ versus $0.8352$, and absolute endpoint change per month is $0.0866$ versus $0.1097$. Their selected-pool similarity medians are $0.1371$ and $0.1911$. These reference contrasts do not establish a stable similarity mechanism. Registry screen-depth summaries are contextual snapshots rather than historical model inputs; aquifer, lithology and land-use classes are unavailable and are not inferred.

\begin{figure}[htbp]
\centering
\includegraphics[width=0.80\linewidth]{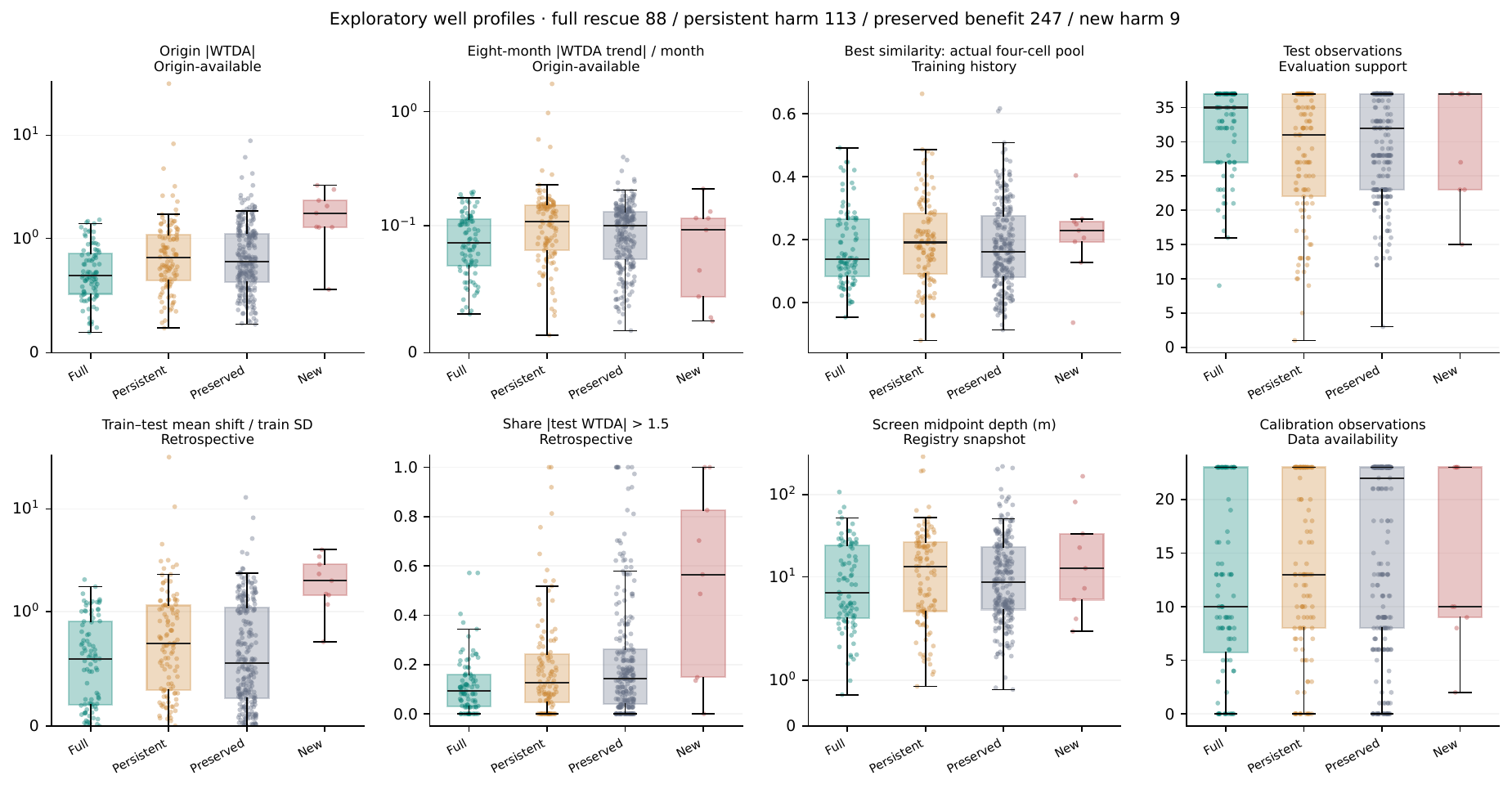}
\caption{Illustrative corrected seed-42 diagnostic. Exploratory distributions for full-rescue (88), persistently harmed (113), preserved-benefit (247) and newly harmed (nine) wells. Panels distinguish forecast-origin inputs, evaluation support, test-truth retrospective diagnostics and BRO registry snapshots. Origin anomaly magnitude, endpoint change, train--test shift and screen depth use symmetric logarithmic axes to retain all outliers. Boxes show median/interquartile range and points retain observations; missing diagnostics remain missing. The small new-harm group and multiple explored features preclude causal or confirmatory interpretation. These distributions do not establish relative predictive importance. Screen metadata are not established historical model inputs.}
\label{fig:apx_success_failure}
\end{figure}

Seed-42 test $\gamma$ has mean $0.46717$, median $0.44791$ and standard deviation $0.21957$. Only $3.16\%$ fall below $0.1$ and $3.97\%$ exceed $0.9$; none are outside $[0,1]$. Under the $\gamma\geq0.5$ descriptive convention, $41.17\%$ favor Always TL and $58.83\%$ favor No TL. Gate fitting uses only 6,562 calibration observations. Nonconstant weights do not establish per-well protection or contextual added value.

\begin{figure}[htbp]
\centering
\includegraphics[width=0.80\linewidth]{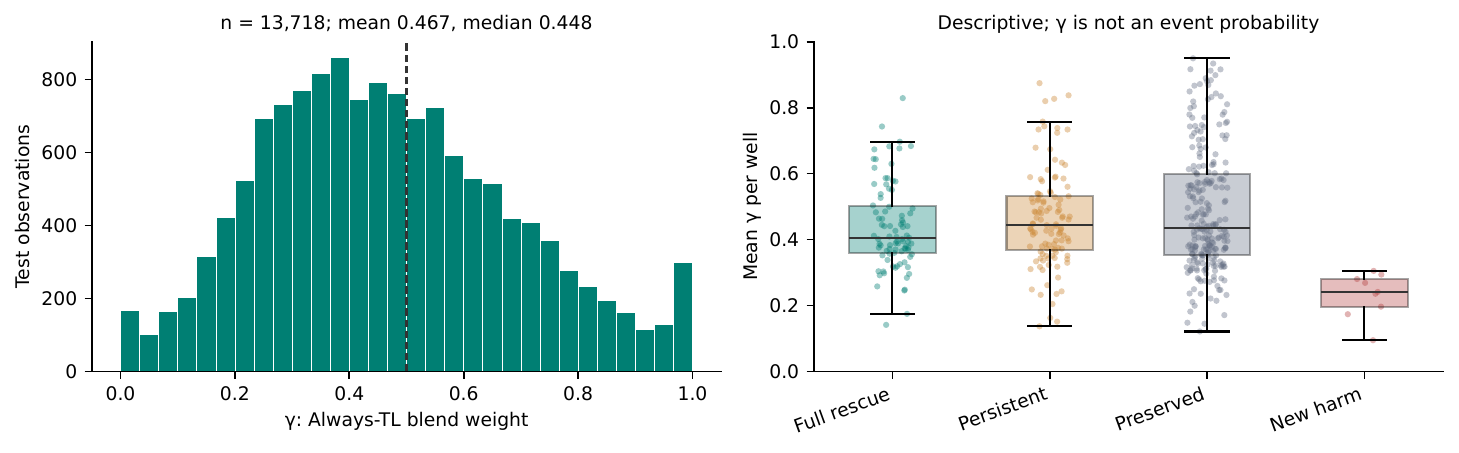}
\caption{Illustrative corrected seed-42 diagnostic. Left: blending weight $\gamma$ over 13,718 corrected test observations. Weights remain within $[0,1]$ without collapse to either parent. Right: mean $\gamma$ per well by outcome group. The $\gamma\geq0.5$ convention describes dominant blend weight, not hard routing, event probability or a cause of group membership. Gate fitting used only 6,562 calibration rows. Absence of collapse does not establish superior harm protection.}
\label{fig:apx_gate_behavior}
\end{figure}

The inference-only group sensitivity uses eight joint permutations of each of nine disjoint groups covering 47 inputs. It averages absolute changes in clipped $\gamma$ within wells and then equally across wells. No estimator is refitted and no test truth enters that calculation. Permutation spread describes randomization variability, not predictive uncertainty. Group size, correlated inputs and unrealistic combinations constrain interpretation; sensitivity is not a training ablation or evidence of a causal mechanism or stronger harm protection than GCB.

\begin{figure}[htbp]
\centering
\includegraphics[width=0.80\linewidth]{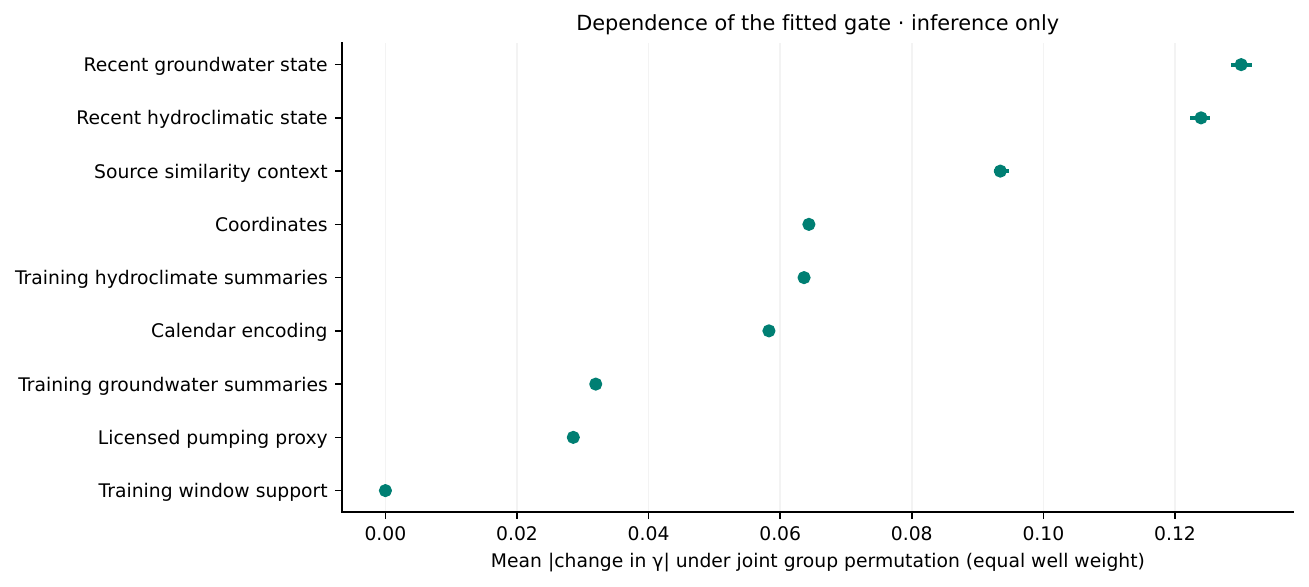}
\caption{Illustrative corrected seed-42 diagnostic. Inference-only sensitivity of the saved \texttt{hgb\_env} gate under joint permutation of nine disjoint input groups covering all 47 features. Dots show mean equal-well-weight absolute change in $\gamma$ over eight permutations; lines show the 5th--95th percentile randomization spread. No estimator is fitted and no test truth enters this calculation. Group size and correlated inputs affect sensitivity, so this is neither conditional importance nor proof of physical mechanism, improved predictive accuracy or added protection over GCB.}
\label{fig:apx_gate_sensitivity}
\end{figure}

\FloatBarrier

\end{document}